\documentclass[runningheads]{llncs}

\usepackage{eccv}

\usepackage{eccvabbrv}

\usepackage{graphicx}
\usepackage{booktabs}

\usepackage[accsupp]{axessibility}  %

\usepackage[pagebackref,breaklinks,colorlinks]{hyperref}

\usepackage{orcidlink}

\newcommand\rurl[1]{%
  \href{https://#1}{\nolinkurl{#1}}%
}

\begin{document}

\title{Ponymation: Learning Articulated 3D Animal Motions from Unlabeled Online Videos}

\titlerunning{Ponymation}

\author{Keqiang Sun\inst{1}$^\ast$\orcidlink{0000-0003-2900-1202} \and
Dor Litvak\inst{2,3}$^\ast$\orcidlink{0009-0004-8720-618X} \and
Yunzhi Zhang\inst{2}\orcidlink{0009-0000-3919-4883} \and
Hongsheng Li\inst{1}\orcidlink{0000-0002-2664-7975} \and \\
Jiajun Wu\inst{2}$^\dagger$\orcidlink{0000-0002-4176-343X} \and
Shangzhe Wu\inst{2}$^\dagger$\orcidlink{0000-0003-1011-5963}}

\authorrunning{K.~Sun et al.}

\institute{
$^{1}$CUHK MMLab \quad $^{2}$Stanford University \quad $^{3}$UT Austin
\\
\url{https://keqiangsun.github.io/projects/ponymation}
}

\maketitle

\def\thefootnote{*}\footnotetext{Equal contribution. $^\dagger$Equal advising.}\def\thefootnote{\arabic{footnote}}

\begin{abstract}

We introduce a new method for learning a generative model of articulated 3D animal motions from raw, unlabeled online videos. Unlike existing approaches for 3D motion synthesis, our model requires \emph{no} pose annotations or parametric shape models for training; it learns purely from a collection of unlabeled web video clips, leveraging semantic correspondences distilled from self-supervised image features. At the core of our method is a video \emph{Photo-Geometric Auto-Encoding} framework that decomposes each training video clip into a set of explicit geometric and photometric representations, including a rest-pose 3D shape, an articulated pose sequence, and texture, with the objective of re-rendering the input video via a differentiable renderer. This decomposition allows us to learn a generative model over the underlying articulated pose sequences akin to a Variational Auto-Encoding (VAE) formulation, but \emph{without} requiring any external pose annotations. At inference time, we can generate new motion sequences by sampling from the learned motion VAE, and create plausible 4D animations of an animal automatically within seconds given a single input image.

\end{abstract}

\section{Introduction}
\label{sec:intro}

We share the planet with a wide variety of lively animals.
Similarly to humans, they navigate and interact with the physical world, demonstrating various sophisticated motion patterns.
In fact, the first film in history, ``The Horse in Motion,'' was a sequence of photographs that captured a galloping horse, created by Eadweard Muybridge in 1887~\cite{Muybridge1887horse}.
Films capture only sequences of 2D projections of 3D animal movements.
Further modeling dynamic animals in 3D is not only useful for numerous mixed reality and content creation applications, but also provides computational tools for biologists to study animal behaviors.

\begin{figure}[t]
\centering
\includegraphics[trim={10pt 0 0 0}, clip, width=0.99\linewidth]{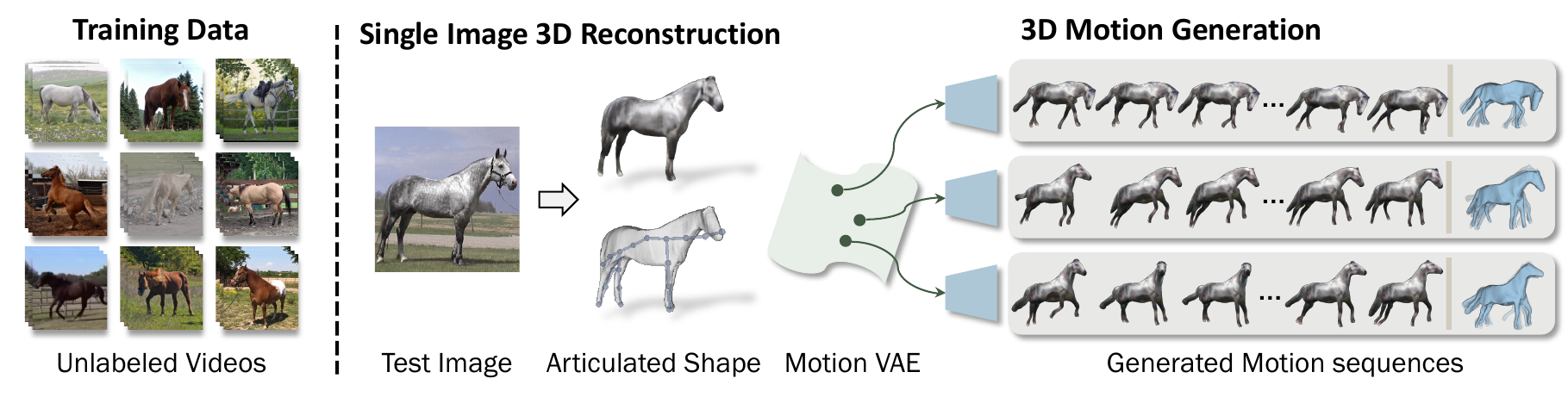}
\vspace{-1em}
\caption{\textbf{Learning 3D Animal Motions from Unlabeled Online Videos.}
Given a collection of monocular videos of an animal category sourced from the Internet as training data, our method learns a \emph{generative} model of the articulated 3D motions together with a monocular 3D reconstruction model, without relying on any shape templates or pose annotations.
At inference time, the model generates new 3D motion sequences and turns a single test image in 4D animations fully automatically.
}
\vspace{-1.5em}
\label{fig:teaser}
\end{figure}

While a lot of efforts have been invested in capturing and modeling 3D human motions using computer vision techniques, significantly less attention has been paid to animals.
Existing learning-based approaches require an extensive amount of 3D scans~\cite{loper2015smpl, pifuSHNMKL19, saito2020pifuhd}, parametric shape models~\cite{Bogo16, Kanazawa_2018_CVPR, Kanazawa_2019_CVPR, zhang2019predicting, piao2021inverting}, multi-view videos~\cite{li2019learning, he2021challencap, gao2022mps}, or geometric annotations, such as 
keypoints~\cite{habibie2017recurrent, henter2020moglow, guo2020action2motion, petrovich21actor, piao2021inverting, petrovich2022temos, starke2022deepphase}, 
as supervision for training.
Collecting large-scale 3D training data involves specialized capture devices and intensive labor, which can only be justified for specific objects, like humans, that are of utmost value in applications.

In this work, we would like to learn a \emph{generative} model of the 3D motions of an animal category, which will allow us to sample new 3D motion sequences and generate 4D animations fully automatically within seconds in a feedforward fashion.
Crucially, unlike existing 3D motion synthesis approaches on human bodies~\cite{henter2020moglow, guo2020action2motion, petrovich21actor, petrovich2022temos, jiang2024motiongpt, xie2023omnicontrol, zhou2023ude}, we do \emph{not} rely on explicit manual supervision for training, such as keypoints or template shapes.
Instead, we propose to learn this 3D motion generative model purely from raw, unlabeled videos sourced from the Internet.
This task is also different from video synthesis methods~\cite{wang2017predrnn, singer2023makeavideo, videoworldsimulators2024} that operate purely on 2D images.
We would like to obtain an \emph{explicit} 3D motion representation, in the form of a 3D mesh and a sequence of articulated 3D poses, which can easily facilitate  downstream applications, including fine-grained controllable 3D animation and motion pattern analysis.

Learning 3D motions from unstructured online video collections is an extremely ill-posed task, as each video clip depicts only a short sequence of 2D projections of a \emph{unique} 4D instance, with unique shape, appearance, motion, and viewpoint that are \emph{not} assumed to reappear in another clip.
This task, therefore, requires registering these unique video clips in a single canonical 3D model to learn a distribution of the underlying 3D motions of the animals.
To address this challenge, we take advantage of recent advancements in self-supervised image representation learning~\cite{caron2021dino}, and distill semantic correspondences across different instances from self-supervised image features produced by a pre-trained DINO-ViT~\cite{caron2021dino}.
Furthermore, we assume a coarse description of the motion skeleton of the animal, \eg ``quadruped,'' which effectively constrains the space of deformation akin to Non-Rigid Structure-from-Motion~\cite{Bregler00nrsfm} and provides a succinct representation for modeling the 3D motion.

Building on top of these insights, we design a video \emph{Photo-Geometric Auto-Encoding} framework for learning 3D motion generative models from unlabeled videos.
At its core is a spatio-temporal transformer that automatically decomposes a video clip into a set of geometric and photometric factors, including a rest-pose 3D mesh, appearance, viewpoint, and a motion latent code that encapsulates the 3D motion of the instance.
This motion latent code is then decoded into a sequence of articulated 3D poses, which are used to animate the rest-pose mesh and re-render a 2D video clip using a differentiable renderer.
This allows us to train the entire model end-to-end like a ``Variational Auto-Encoder'' (VAE) over the space of articulated 3D motions, using only 2D image reconstruction losses on the RGB frames, DINO features, and object masks, with pseudo-ground-truth masks obtained from off-the-shelf detectors~\cite{kirillov2019pointrend}.

At inference time, we can generate new 3D motion sequences by sampling from the motion VAE latent space.
If further given a single image of an animal, our model can reconstruct its articulated 3D shape and appearance in a feed-forward fashion, and generate 4D animations fully automatically within seconds.

To summarize, this paper makes several contributions:
\vspace{-0.5em}
\begin{itemize}
    \item We propose a new method for learning a \emph{generative} model of articulated 3D animal motions from \emph{unlabeled} Internet videos, without any shape templates or pose annotations;
    \item We design a spatio-temporal transformer architecture that effectively extracts motion information from input video clips into a latent VAE;
    \item At inference time, the model generates diverse 3D motion sequences and turns a single image into 4D animations automatically in seconds;
\end{itemize}

\vspace{-1.5em}
\section{Related Work}

\vspace{-1em}
\subsubsection{Learning 3D Animals from Image Collections.}
While modeling dynamic 3D objects traditionally requires motion capture markers or simultaneous multi-view captures~\cite{hartley04multiple, deAguiar08, Debevec12}, recent learning-based approaches have demonstrated the possibility of learning 3D deformable models simply from raw single-view image collections~\cite{kanazawa18cmr, wu20unsupervised, Li2020umr, yao2022lassie, wu2023magicpony, yao2024artic3d, liu2024lepard, stathopoulos2023learning}.
Most of these methods require additional geometric supervision besides object masks for training, such as keypoint~\cite{kanazawa18cmr, li20online} and viewpoint annotations~\cite{sitzmann2019srns, Niemeyer20DVR, duggal2022tars3D}, template shapes~\cite{Goel20ucmr, kulkarni2020articulation, kokkinos2021point}, semantic correspondences~\cite{Li2020umr, yao2022lassie, wu2023magicpony, yao2023hi, jakab24farm3d}, and strong geometric assumptions like symmetries~\cite{wu20unsupervised, wu2021derender, wu2022casa} and viewpoint distributions~\cite{nguyen2019hologan, Schwarz2020graf, Niemeyer2020GIRAFFE, chan2021piGAN, Chan2022eg3d, sun2022cgof, sun2022cgof++}.
Among these, MagicPony~\cite{wu2023magicpony} demonstrates impressive results in learning articulated 3D animals, such as horses, using only single-view images with object masks and self-supervised image features as training supervision.
However, it reconstructs static images individually, ignoring the dynamic motions of the underlying 3D animals underneath those images.
In this work, we focus on learning a generative model of 3D animal motions from videos instead of independent images.

\vspace{-1.5em}
\subsubsection{Deformable Shapes from Monocular Videos.}
Reconstructing deformable shapes from monocular videos is a long-standing problem in computer vision.
Early approaches with Non-Rigid Structure from Motion (NRSfM) reconstruct deformable shapes from 2D correspondences, by incorporating heavy constraints on the motion patterns~\cite{Bregler00nrsfm, Xiao04nrsfm, Akhter08nrsfm, Dai12, Cashman12dolphin}.
DynamicFusion~\cite{newcombe15dynamicfusion} further integrates additional depth information from depth sensors.
NRSfM pipelines have recently been revived with neural representations.
In particular, LASR~\cite{yang21lasr} and its follow-ups~\cite{yang2021viser, wu2022casa, yang2022banmo, yang2023rac} optimize deformable 3D shapes over a small set of monocular videos, leveraging 2D optical flows in a heavily engineered optimization procedure.
DOVE~\cite{wu2021dove} proposes a learning-based framework that learns a category-specific single-image 3D reconstruction model from a monocular video collection.
Despite using video data for training, none of these approaches explicitly model the generative distribution of temporal motions of the objects.

\vspace{-1em}
\subsubsection{Motion Analysis and Synthesis.}
Modeling motion patterns of dynamic objects has important applications for both behavior analysis and content generation, and is instrumental to our visual perception system~\cite{Badler1975}.
Computational techniques have been used for decades to study and synthesize human motions~\cite{Badler1993, Ormoneit2005, Urtasun2007}.
In particular, recent works have explored learning generative models for 3D human motions~\cite{Lin2018HumanMM, Ahn2018, minderer2019unsupervised, henter2020moglow, guo2020action2motion, petrovich21actor, petrovich2022temos, starke2022deepphase, kapon2023mas}, leveraging parametric human shape models, like SMPL~\cite{loper2015smpl}, and large-scale human pose annotations~\cite{Ionescu2014, Andriluka2014}.
In comparison, much less effort is invested in modeling animal motions.
Huang \etal~\cite{huang2021hierarchical} proposes a hierarchical motion learning framework for animals, but requires costly motion capture data and hardly generalizes to animals in the wild.
To sidestep the collection of 3D data, BKinD~\cite{bkind2021} introduces a self-supervised method for discovering and tracking keypoints from videos, but is limited to a 2D representation.
Such 2D keypoints could be lifted to 3D~\cite{sun2022bkind, kapon2023mas},
but this requires multi-view videos or ground-truth keypoints for training.
Unlike these prior works, our motion learning framework does not require any pose annotations or multi-view videos for training, and is trained simply using raw monocular online videos.
Recent success of image diffusion models has also led to promising generic 4D generation models~\cite{ren2023dreamgaussian4d, zheng2023unified4d, ling2023ayg, zhao2023animate124, bahmani20234dfy}.
However, the 3D motions generated by these models are still very limited in terms of quality and diversity, as shown in the comparisons in \Cref{sec:motion_eval}.

\vspace{-0.5em}
\section{Method}
\begin{figure*}[t]
\centering
\includegraphics[trim={10pt 0 0 0}, clip, width=0.99\linewidth]{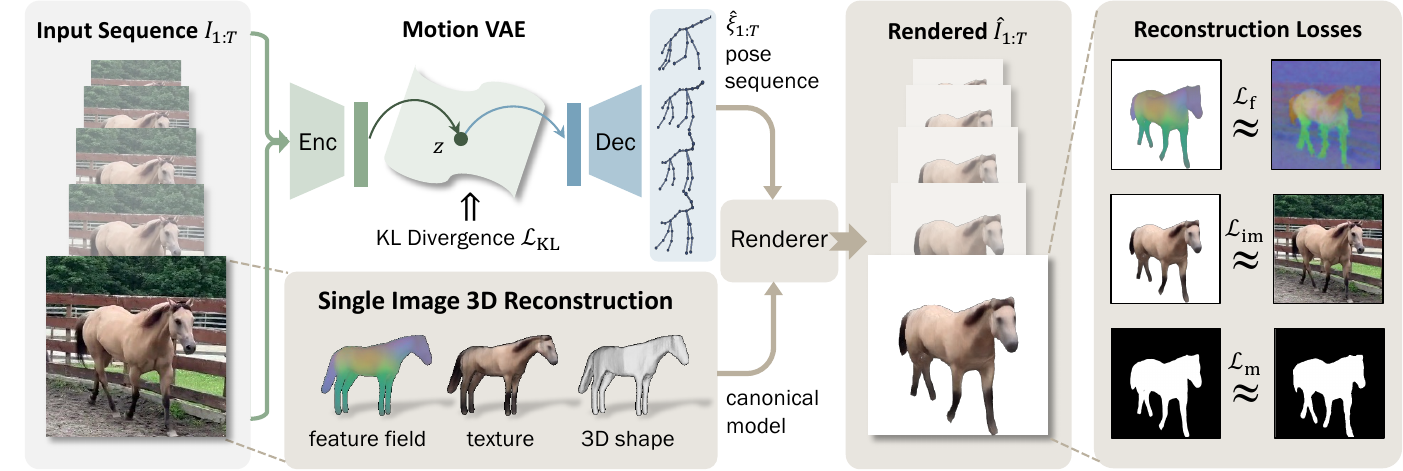}
\vspace{-0.5em}
\caption{\textbf{Training Pipeline.}
Our method learns a generative model of articulated 3D motion sequences
from a collection of unlabeled monocular videos.
During training,
the model encodes an input video sequence $I_{1:T}$ into a latent code $z$ in the motion VAE, and decodes from it a sequence of articulated 3D poses $\hat{\xi}_{1:T}$.
This pose sequence is used animate the reconstructed 3D shape, allowing the full pipeline to be trained simply using image reconstruction losses with unsupervised image features and object masks obtained from off-the-shelf models, without any external pose annotations.
}%
\vspace{-1em}
\label{fig:method}
\end{figure*}

\vspace{-0.5em}
Given a collection of raw video clips of an animal category, such as horses, our goal is to learn a generative model of its articulated 3D motions.
This allows us to sample 3D motion sequences from a learned latent space, and generate 4D animations of a new animal instance automatically given only a single 2D image at test time.
We train this model simply on raw online videos without relying on any external pose annotations.
To do so, we design a video photo-geometric auto-encoding framework that decomposes each training video clip into a rest-pose 3D mesh, appearance, camera viewpoint as well as a sequence articulated 3D poses.
This allows us to learn a \emph{generative} model over the underlying articulated 3D pose sequences akin to a motion ``Variational Auto-Encoder'', but simply using the objective of re-rendering the input frames with a differentiable renderer.
\Cref{fig:method} gives an overview of the training pipeline.

\vspace{-1em}
\subsection{Modeling Articulated 3D Animal Motions}
\label{sec:modeling_articulated}
\vspace{-0.3em}
Each video clip records a 2D image sequence $\{I_t\}_{t=1}^T$ of the underlying 3D animal motion from one camera trajectory.
Since the dataset is obtained from casually-recorded Internet videos, these training clips have diverse unique motion sequences.
In order to learn the distribution of the underlying animal motions from such unstructured video collections, we first need to devise a 3D representation that registers these dynamic 2D sequences onto a canonical 3D model, factoring out the 3D motion of each video instance.

Drawing inspiration from prior work on 3D human motion synthesis~\cite{loper2015smpl, jiang2024motiongpt, zhou2023ude, xie2023omnicontrol},
we leverage a category-specific skinned model to represent the deformable 3D shape of the animals, and further learn the motion distribution over the articulations of its underlying skeleton.
To this end, we follow MagicPony~\cite{wu2023magicpony} and assume a coarse description of the skeleton, \eg ``quadruped''.

Specifically, we represent the category-specific base 3D shape using a Signed Distance Function (SDF) parametrized by a coordinate Multi-Layer Perceptron (MLP), and extract an explicit mesh on the fly using Differentiable Marching Tetrahedron (DMTet)~\cite{shen2021dmtet}.
Let $V_\text{base} \in \mathbb{R}^{K \times 3}$ denote the list of $K$ vertices, and the triangle faces are given by the triplets $F \subset \{1,\dots,K\}^3$.
To model the slight shape variation of each animal instance in the canonical pose, we further learn an image-conditioned deformation field $f_{\Delta V}$ parametrized by another MLP that predicts small deformations of each vertex $\Delta V_{\text{ins},i} = f_{\Delta V}(V_{\text{base},i}, \phi)$, where $\phi = f_\phi(I)$ is a feature vector obtained from an image $I$ using a pre-trained DINO-ViT~\cite{caron2021dino}, and $i \in \{1, \cdots, K\}$ denotes the vertex index.
Both base shape $V_\text{base}$ and the instance deformation $\Delta V_{\text{ins}}$ are enforced to be bilaterally symmetric about $yz$-plane by mirroring the query locations in the underlying MLPs.

To account for the temporal motions driven by the underlying bone structure, we then instantiate a quadrupedal skeleton in this instance shape using a simple heuristic:
a chain of bones going through the two farthest end points along $z$-axis, and four legs branching out from the body bone to the lowest point in each $xz$-quadrant.
The motion sequence is thus parametrized by a sequence of articulated poses $\xi = \{\xi_t\}_{t=1}^T$, where each pose $\xi_t$ at a timestamp $t$ consists of a rigid pose $\xi_{t,1} \in SE(3)$ \wrt an identity camera pose and the rotation $\xi_{t,b} \in SO(3)$ of each bone $b = 2, ..., B$ in the skeleton.
These articulated poses are applied to the instance mesh $V_{\text{ins}}$ to obtain the final posed shape sequence using the widely-used linear blend skinning $g(V_{\text{ins}}, \xi_t)$~\cite{loper2015smpl}.
More details are included in the supplementary material.

The appearance of the instance is modeled using a texture field parametrized by an MLP $f_\text{a}(\mathbf{x}, \phi) \in [0,1]^3$ where $\mathbf{x}$ is a 3D location.
We then render the posed mesh sequence into a sequence of RGB images using deferred mesh rendering~\cite{wu2023magicpony}, querying $f_\text{a}$ at the corresponding 3D locations of the pixels after rasterization.

In the following, we explain the learning formulation to learn the individual components, including $V_\text{base}$, $f_{\Delta V}$, $f_\text{a}$, and most importantly, a generative model $f_\xi$ over the motion sequences $\xi$, purely from an unstructured video collection without external pose annotations.

\vspace{-0.5em}
\subsection{Video Photo-Geometric Auto-Encoding}
\label{sec:video_audoencoding}
\vspace{-0.5em}
Unlike human motion synthesis, we do not have access to large-scale, high-quality 3D captures or pose annotations for most animal species. Hence, we must instead learn from raw Internet videos, which poses significant challenges.
To this end, we design a video \emph{Photo-Geometric Auto-Encoding} framework that deconstructs each training clip into the explicit photometric and geometric factors described in \Cref{sec:modeling_articulated}, and train the entire pipeline using the objective of re-rendering the video.
At the center of this video auto-encoding pipeline is a generative model of articulated motion sequences, akin to a ``Variational Auto-Encoder'' (VAE), but learned purely from raw RGB frames.
This is very different from simply training a conventional VAE directly in the pose sequence space, which would require explicit pose annotations in the first place.

\vspace{-1em}
\subsubsection{Video Encoding.}
To predict the instance shape deformation $\Delta V_\text{ins}$ and appearance of the object, we extract a feature vector $\phi_t$ for each frame of the video using a pre-trained DINO-ViT~\cite{caron2021dino} with frozen weights, as mentioned previously.
We assume the instance shape and appearance remain the same throughout the video, and hence take the average image features across all frames, denoted as $\bar{\phi}$, when querying the MLPs, $f_{\Delta V}$ and $f_\text{a}$.

In order to extract the motion information more effectively from the input video clip, we design a pair of spatial and temporal transformer-based motion encoders, $E_\text{s}$ and $E_\text{t}$, that aggregate a set of bone-specific local features first spatially across each frame and then temporally across the entire sequence, eventually obtaining the distribution parameters $\hat{\mu}$ and $\hat{\Sigma}$ of the motion latent VAE.

Specifically, given each frame $I_t$ in the input clip, we first construct a bone-specific feature descriptor $\nu_{t,b} = (\phi_t, \Phi_t(\mathbf{u}_{t,b}), b, \mathbf{J}_b, \mathbf{u}_{t,b})$ for each bone $b = 2, ..., B$ and each timestamp $t$.
Here, $\phi_t$ denotes the same global image feature as before.
$\mathbf{J}_b$ denotes the 3D location of the center of the bone $b$ at rest-pose, which projects to the pixel location $\mathbf{u}_{t,b}$ in the image space, given the rigid pose $\hat{\xi}_{t,1}$ predicted separately.
In addition to the global feature $\phi_t$, we also sample an auxiliary bone-specific local feature vector $\Phi_t(\mathbf{u}_{t,b})$ from the DINO-ViT key token map $\Phi_t$ at the projected pixel location $\mathbf{u}_{t,b}$.

The spatial transformer encoder $E_\text{s}$ then fuses these bone-specific feature descriptors $\{\nu_{t,b}\}_{b=2}^B$ into a single feature vector $\nu_{t,*}$ summarizing the articulated pose of the animal in each frame $t$:
\begin{equation}
    \nu_{t,*} = E_\text{s} (\nu_{t,2}, \cdots, \nu_{t,B}).
\end{equation}
In practice, we prepend a learnable token to the list of descriptors, and take the first output token of the transformer as the pose feature $\nu_{t,*}$.
We call this $E_\text{s}$ a \emph{spatial} transformer as it extracts the spatial geometric features in each input frame that capture the pose information, conditioned on the given skeleton.

Next, we design a second \emph{temporal} transformer encoder $E_\text{t}$, inspired by \cite{petrovich21actor}, which operates along the temporal dimension and maps the entire sequence of pose features $\{\nu_{t,*}\}_{t=1}^T$ into the motion latent space.
Similarly to the $E_\text{s}$, $E_\text{t}$ fuses the pose feature sequence to predict the VAE distribution parameters:
\begin{equation}
    (\hat{\mu}, \hat{\Sigma}) = E_\text{t} (\nu_{1,*}, \cdots, \nu_{T,*}).
\end{equation}
Using the reparametrization trick~\cite{kingma2013auto}, we then sample a latent code from the Gaussian distribution $z\sim \mathcal{N}(\hat{\mu}, \hat{\Sigma})$, which will be decoded into a sequence of articulated poses $\{\hat{\xi}_{t}\}_{t=1}^T$ characterizing the 3D motion of the animal in the clip.

\vspace{-1em}
\subsubsection{Motion Decoding.}
Symmetric to the motion encoders, the motion decoder also consists of a temporal decoder $D_\text{t}$ that first decodes $z$ into a sequence of pose features $\{z_t\}_{t=1}^T$, and a spatial decoder $D_\text{s}$ that further decodes each pose feature $z_t$ to a set of bone rotations $\{\hat{\xi}_{t,b}\}_{b=2}^B$.

Specifically, we query the temporal transformer decoder $D_\text{t}$ with a sequence of timestamps $\mathcal{T}$, and use $z$ as both the key token and the value token to obtain a sequence of pose features:
\begin{equation}
    (z_1, \cdots, z_T) = D_\text{t} (\mathcal{T}, z), \quad \mathcal{T} = (1, \cdots, T).
\end{equation}
Similarly, given each pose feature $z_t$, we then query the spatial transformer decoder $D_\text{s}$ with a sequence of bone indices $\mathcal{B}$ to produce the bone rotations:
\begin{equation}
    (\hat{\xi}_{t, 2}, \cdots, \hat{\xi}_{t, B}) = D_\text{s} (\mathcal{B}, z_t), \quad \mathcal{B} = (2, \cdots, B).
\end{equation}
In practice, the rigid pose $\hat{\xi}_{t,1}$ is predicted by a separate network and is not modeled by this motion VAE, since it is entangled with arbitrary camera motions that are difficult to disentangle in dynamic scenes.

We then deform the predicted instance mesh $\hat{V}_\text{ins}$ using these articulated pose sequence $\{\hat{\xi}_t\}_{t=1}^T$ with the skinning equation $\hat{V}_t = g (\hat{V}_\text{ins}, \hat{\xi}_t)$, and render the RGB frames $\{\hat{I}_t\}_{t=1}^T$ and masks $\{\hat{M}_t\}_{t=1}^T$ using a differentiable renderer~\cite{Laine2020diffrast}.

\vspace{-1em}
\subsection{Learning Formulation}

\subsubsection{Video Re-rendering Losses.}
We train the entire model by minimizing the reconstruction losses on the object masks $\hat{M}_t$ and RGB frames $\hat{I}_t$:
\begin{equation}
    \mathcal{L}_{\text{m},t} = \|\hat{M}_t - M_t\|_2^2 + \lambda_\text{dt} \|\hat{M}_t \odot \texttt{dt}(M_t)\|_1,
    \quad
    \mathcal{L}_{\text{im},t} = \|\tilde{M}_t \odot (\hat{I}_t - I_t)\|_1,
\end{equation}
where distance transform $\texttt{dt}(\cdot)$ is used in the second term of the mask loss with a weight $\lambda_\text{dt}$ for more effective gradients~\cite{kanazawa18cmr, wu2021derender, wu2023magicpony}, and $\odot$ denotes the Hadamard product.
The RGB loss is only computed inside the intersection of the predicted and ground-truth masks $\tilde{M}_t = \hat{M}_t \odot M_t$.
To exploit the temporal consistency of the motion in the videos, we further enforce a temporal smoothness constraint between the predicted poses $\hat{\xi}_t$ of consecutive frames:
$\mathcal{R}_\text{temp} = \sum_{t=2}^{T} \|\hat{\xi}_t - \hat{\xi}_{t-1}\|_2^2 $.
We also inherit the multi-hypothesis viewpoint prediction mechanism with the hypothesis loss $\mathcal{L}_\text{hyp}$ and the shape regularizers
$
\mathcal{R}_\text{shape} = \lambda_\text{Eik} \mathcal{R}_\text{Eik} +
\lambda_\text{art} \mathcal{R}_\text{art} +
\lambda_\text{def} \mathcal{R}_\text{def}
$~\cite{wu2023magicpony} with balancing weights $\lambda$'s, which include the Eikonal constraint $\mathcal{R}_\text{Eik}$ on the SDF MLP for the base shape, and magnitude regularizers $\mathcal{R}_\text{art}$ on the bone rotations $\hat{\xi}_{2:B}$ and $\mathcal{R}_\text{def}$ on the vertex deformations $\Delta V_\text{ins}$.

\vspace{-1em}
\subsubsection{Semantic Correspondences.}
Instead of relying on external pose annotations or prior shape models to learn the 3D model from monocular videos, we seek a much cheaper alternative solution for establishing correspondences across different instances.
We distill semantic correspondences from self-supervised image features, such as DINO~\cite{caron2021dino}.
As shown in prior work~\cite{amir2021deep, wu2023magicpony, yao2022lassie}, after a simple PCA reduction, these image features reveal robust part-level correspondences across different instances with varying poses and appearance.
To exploit these correspondences, we additionally optimize a feature field in the canonical space using a coordinate MLP $\psi(\mathbf{x}) \in \mathbb{R}^D$, which is rendered into an 2D feature image $\hat{\Phi}_t \in \mathbb{R}^{D \times H \times W}$ given the posed mesh $\hat{V}_t$, with the same procedure as rendering the appearance of the object described above.
We then encourage this rendered feature map $\hat{\Phi}_t$ to match the feature map $\Phi'_t$ pre-extracted from the input frame $I_t$ using DINO-ViT with PCA reduction:
$
    \mathcal{L}_{\text{feat},t} = \|\tilde{M}_t \odot (\hat{\Phi}_t - \Phi'_t)\|_2^2.
$
Intuitively, this enforces the model to establish correspondences across all training video instances through the same canonical feature field, hence disentangling the shape and pose in each monocular frame.

\vspace{-1em}
\subsubsection{Motion VAE.}
Similarly to the conventional VAE, we also minimize the Kullback–Leibler (KL) divergence between the learned motion latent distribution and a standard Gaussian distribution:
\begin{equation}
\mathcal{L}_\text{KL} = \sum_i -\frac{1}{2}\left(\log \sigma_i - \sigma_i - \mu_i^2 + 1\right),
\vspace{-0.5em}
\end{equation}
where $\mu_i$ and $\sigma_i^2$ are elements of the predicted distribution parameters $\hat{\mu}$ and $\hat{\Sigma}$.

\vspace{-1em}
\subsubsection{Training Schedule.}
\label{sec:training_schedule}
As learning 3D articulated motions from unstructured video clips without labels is extremely ill-posed, we devise a two-stage schedule for robust and efficient training.
In the \emph{first} stage, we pre-train the monocular 3D reconstruction model using a single-image pose predictor $\tilde{\xi}_t = f_\xi^\text{sin}(\phi_t)$.
Inspired by but unlike \cite{wu2023magicpony}, we train this model to re-render entire video clips with the temporal smoothness constraint $\mathcal{R}_\text{temp}$ and temporal feature averaging $\bar{\phi}$, rather than independent images.
The total loss in the first stage is given by:
\begin{equation}
\vspace{-0.5em}
\label{eq:video_recon_loss}
    \mathcal{L}_\text{vid} =
    \sum_{t=1}^T \left(
        \mathcal{L}_{\text{recon},t} +
        \lambda_\text{h} \mathcal{L}_{\text{hyp},t} +
        \lambda_\text{s} \mathcal{R}_{\text{shape},t}
    \right) +
    \lambda_\text{t} \mathcal{R}_\text{temp},
\end{equation}
where
$
\mathcal{L}_{\text{recon},t} =
    \mathcal{L}_{\text{im},t} +
    \lambda_\text{m} \mathcal{L}_{\text{m},t} +
    \lambda_\text{f} \mathcal{L}_{\text{feat},t}
$
summarizes the reconstruction losses on each frame.
After this stage, we obtain an accurate monocular 3D reconstruction model, which outperforms the baseline~\cite{wu2023magicpony} as shown in \Cref{tab:recon}, largely owing to the training on videos instead of independent images.
More importantly, the model has now learned a reasonable space of articulated poses, on top of which learning a motion generative model is much more efficient.

In the \emph{second} stage, we replace the monocular pose predictor $f_\xi^\text{sin}$ with the spatio-temporal transformer-based motion VAE $f_\xi$ detailed in \Cref{sec:video_audoencoding}, which encodes the entire video clip and generates the entire sequence of articulated poses at once.
Empirically, training the motion VAE from scratch with an expensive rendering step in the loop is inefficient.
To facilitate training efficiency, we recycle pose predictions $\tilde{\xi}_t$ from the first stage to guide the predictions of the VAE decoder $\hat{\xi}_t$ using a teacher loss $\mathcal{L}_\text{teacher} = \sum_{t=1}^T \| \hat{\xi}_t - \tilde{\xi}_t \|_2^2$.
The final training objective for the second stage is thus:
\begin{equation}
    \label{eqn:final}
    \mathcal{L} = \mathcal{L}_\text{vid} + \lambda_\text{KL} \mathcal{L}_\text{KL} + \lambda_\text{teacher} \mathcal{L}_\text{teacher}.
\end{equation}

\vspace{-1.5em}
\subsubsection{3D Motion Generation.}
\begin{figure}[th!]
\centering
\includegraphics[trim={8pt 0 5pt 0}, clip, width=\linewidth]{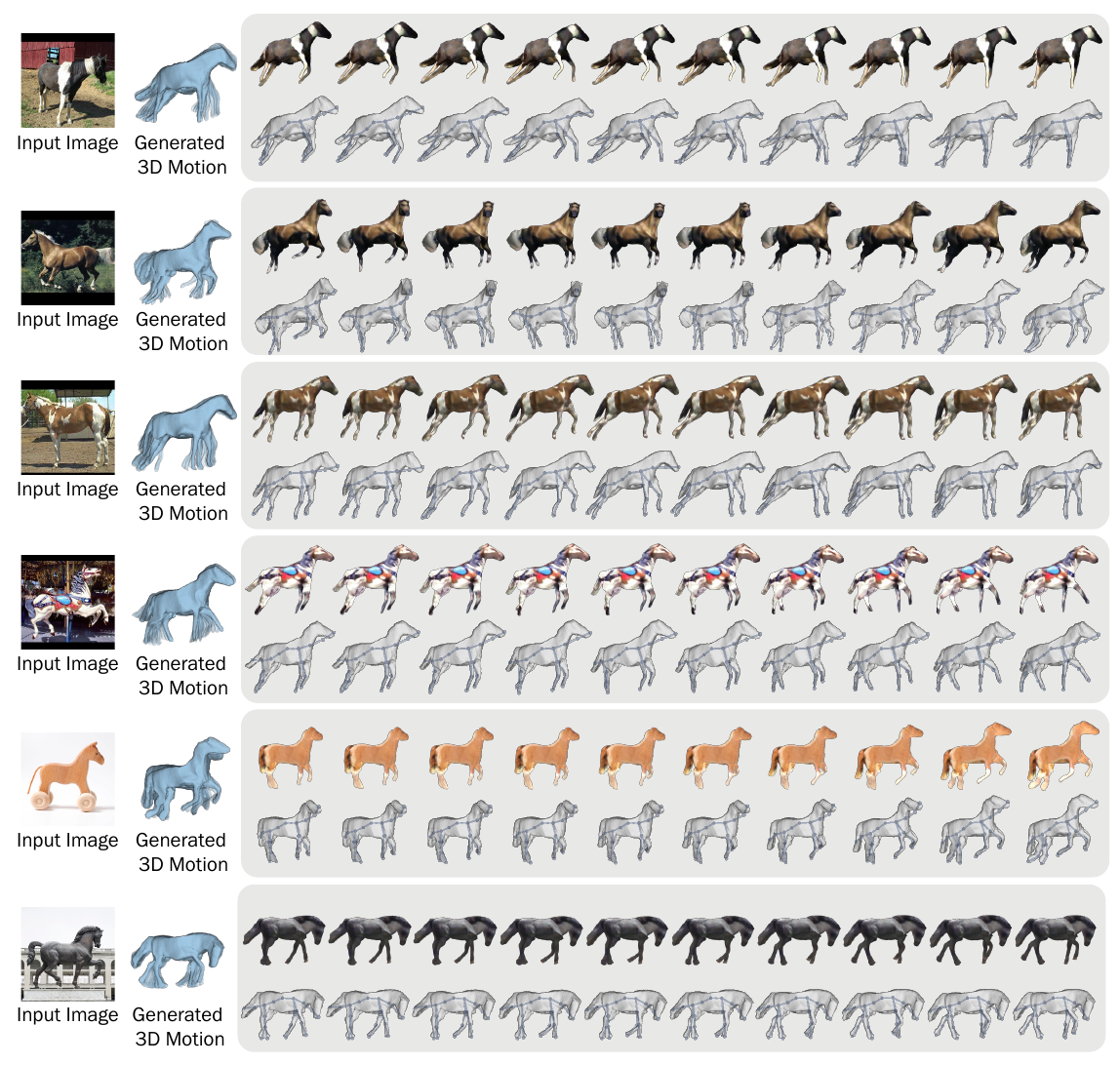}
\vspace{-2em}
\caption{\textbf{3D Motion Generation and Animation.}
During test time, our model generates plausible 3D motion sequences by sampling from the learned motion VAE.
It can also reconstruct articulated 3D shapes from a single 2D image in feed-forward fashion, and generate 4D animations fully automatically within seconds.
Within each gray box on the right, the first row shows textured animation, and the second row visualizes the corresponding 3D shapes with the generated bone articulations.
}%
\vspace{-1em}
\label{fig:motion}
\end{figure}

During inference time, we can generate diverse 3D motion sequences by sampling from the learned motion VAE latent space.
Furthermore, when given a single 2D image of a new animal instance unseen at training, our model can reconstruct its 3D shape and appearance in a feed-forward manner, and generate 4D animations fully automatically within a few seconds, as illustrated in \Cref{fig:motion}.

\begin{table}[t!]
\small
\setlength{\tabcolsep}{12pt}
\centering
\vspace{-0.5em}
\caption{\label{tab:dataset} Statistics of the \emph{AnimalMotion} Dataset.
We collect a new animal video dataset containing a total of $82.6$k frames for 4 different animal species.}
\begin{tabular}{lccc}
\toprule
\multicolumn{1}{c}{Category}    & \# Sequences & Total Length    & \# Frames \\
\midrule
Horse       &640    & 28'09'' &50,682\\
Zebra       &47	    & 5'27'' &9,822\\
Giraffe     &60	    & 4'52'' &8,768\\
Cow         &69     & 7'25'' &13,359\\
\midrule
Total       &816    & 45'54'' &82,631\\
\bottomrule
\end{tabular}
\vspace{-1em}
\end{table}

\vspace{-1em}
\section{Experiments}
\vspace{-0.5em}
\subsection{Experimental Setup}
\vspace{-0.5em}
\subsubsection{Datasets.} 
To train our model, we collected an \textbf{AnimalMotion} dataset consisting of video clips of several quadruped animal categories extracted from the Internet.
The statistics of the dataset are summarized in \Cref{tab:dataset}.
As pre-processing, we first detect and segment the animal instances in the videos using the off-the-shelf segmentation model of PointRend~\cite{kirillov2019pointrend}.
To remove occlusion between different instances, we calculated the extent of mask overlap in each frame and exclude crops where two or more masks overlap with each other.
We further apply a smoothing kernel to the sequence of bounding boxes to avoid jittering.
The non-occluded instances are then cropped and resized to $256 \times 256$.
The original videos are all at 30fps.
To ensure sufficient motion in each sequence, we remove frames with minimal motion, measured by the magnitude of optical flows within the instance mask estimated from RAFT~\cite{teed2020raft}.
To conduct quantitative evaluations and comparisons, we also use PASCAL VOC~\cite{everingham2015pascal} which contains $108$ images of horses,
and APT-36K~\cite{yang2022apt} which contains $81$ video clips of horses, each consisting of $15$ frames.
Both datasets provide 2D keypoint annotations for each animal in the image, allowing us to evaluate the geometric accuracy of the reconstructed shapes and generated motions.

\vspace{-1em}
\subsubsection{Implementation Details.}
The encoders and decoders of the motion VAE model ($E_\text{s}$, $E_\text{t}$, $D_\text{s}$, $D_\text{t}$) from \Cref{sec:video_audoencoding} are implemented as stacked transformers~\cite{transformer} with 4 transformer blocks and a latent dimension of 256.
We use a sinusoidal function for positional encoding following \cite{petrovich21actor}.
For the remaining architectures, we base our implementation on top of \cite{wu2023magicpony}.
We train the model for $120$ epochs for the first stage, which takes roughly $10$ hours on $8$ A6000 GPUs, and another $180$ epochs for the second stage, which takes another $48$ hours.
We use a sequence length of $T=10$ for training.
During inference, we can generate longer sequences by connecting multiple samples and optimizing transition latent codes for smooth interpolation.
For visualization, following prior work~\cite{wu2023magicpony}, we finetune (only) the appearance network $f_\text{a}$ for $100$ iterations on each test image, taking less than 10 seconds, as the model struggles to predict detailed texture in a single feedforward pass.
More details are included in the sup.~mat.

\vspace{-0.5em}
\subsection{3D Motion Generation}

\vspace{-0.5em}
\subsubsection{Qualitative Results.}
\begin{figure}[t!]
\centering
\includegraphics[trim={2mm 0 0 0}, clip, width=\linewidth]{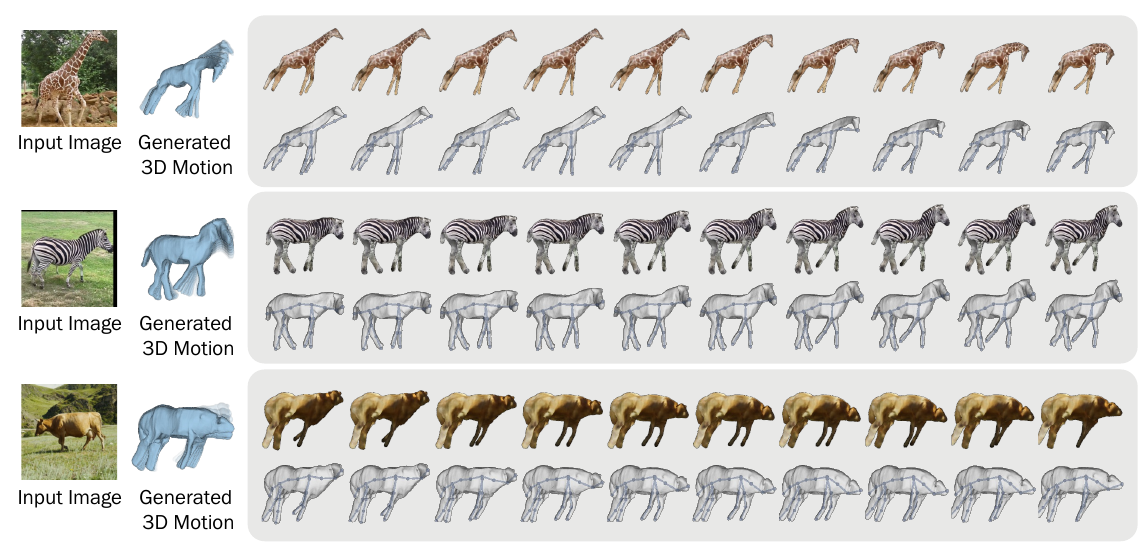}
\vspace{-2em}
\caption{\textbf{3D Motion Generation Results on More Species.}
Our method can be trained on various animal species, such as corws, zebras, and giraffes illustrated here.
The model learns to generate 3D motions
and generate plausible motion sequences specific to the animal species, such as the generated neck motion in the first example which is more common in giraffes than others. 
}%
\vspace{-0.5em}
\label{fig:other_categories}
\end{figure}

After training, we can generate 3D motion sequences by sampling the motion latent space VAE, and render 4D animations with the textured mesh reconstructed from a single 2D image, as shown in \Cref{fig:motion}.
It also generalizes to horse-like artifacts, such as carousel horses, which the model has never seen during training.
The model can be trained on a wide range of animal species besides horses, including giraffes, zebras and cows, capturing category-specific prior distributions of 3D motions, as shown in \Cref{fig:other_categories}.
Because the datasets for these categories are limited in size and diversity, as in \cite{wu2023magicpony}, in the first stage of the training, we fine-tune from the model trained on horses.
Additional animation results are provided in the supplementary video.

\begin{figure}[t!]
\centering
\includegraphics[trim={0mm 0 0 0}, clip, width=\linewidth]{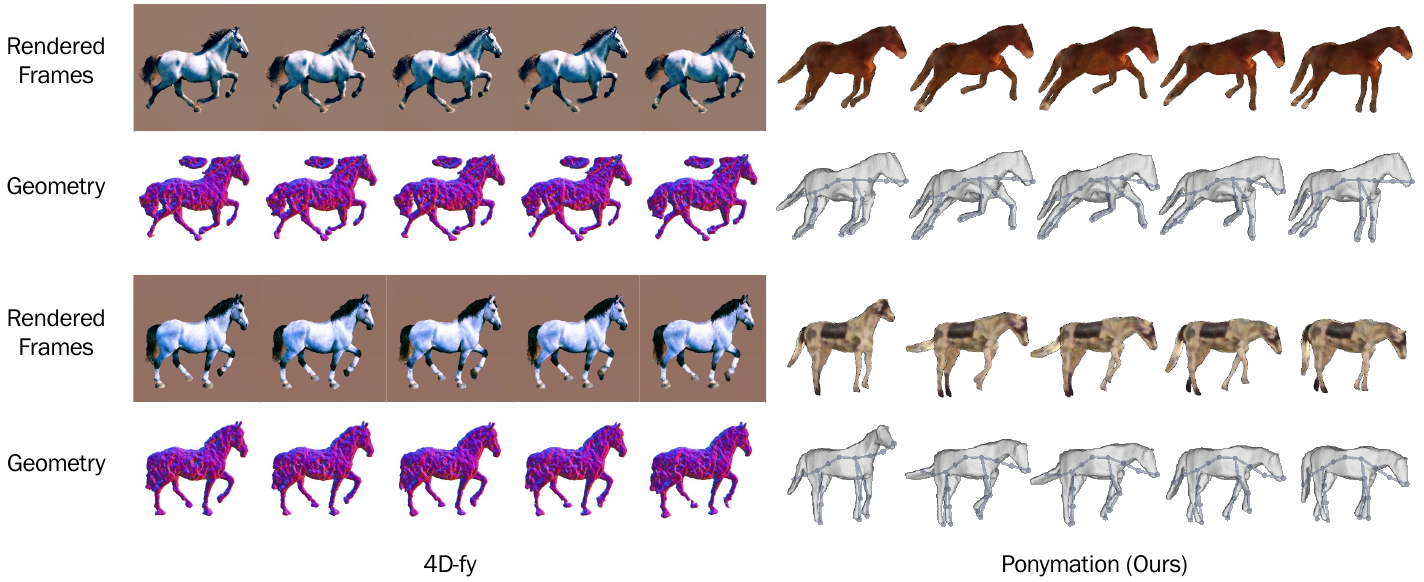}
\vspace{-2em}
\caption{\textbf{4D Generation Comparisons.}
We compare with 4D-fy~\cite{bahmani20234dfy}, a recent text-to-4D generation method distilling from 2D diffusion.
Despite heavy prompt engineering and a lengthy training time (12 hours), 4D-fy still fails to produce noticeable motion, whereas our model generates diverse motion sequences in a feed-forward pass within a few seconds, with much better 3D geometry.
}%
\vspace{-1em}
\label{fig:compare_4dfy}
\end{figure}

\begin{table}[t!]
\setlength{\tabcolsep}{0.2cm}
\centering
\footnotesize
\caption{
Quantitative Comparison with State-of-the-Art Motion Generative Models.}
\vspace{-0.5em}
\begin{tabular}{lcc}
\toprule
 & Motion Strength & User Preference \\
\midrule 
4D-fy~\cite{bahmani20234dfy} & 0.29 & 112 (17.0\%)\\
Ponymation (Ours) & \textbf{4.66} &\textbf{548} (\textbf{83.0\%}) \\

\bottomrule
\end{tabular}
\vspace{-1.2em}
\label{tab:cmp_4dfy}
\end{table}

\vspace{-1em}
\subsubsection{Comparison with Existing Methods.}
\label{sec:motion_eval}
Our method is the first to learn a generative model of 3D animal motions from raw videos without pose annotations or prior shape models.
We compare with one of the most recent 4D generative models, 4D-fy\cite{bahmani20234dfy}, which has publicly released code.
Specifically, we provide the model with a list of prompts, which are enriched by ChatGPT~\cite{chatgpt} from a list of basic prompts describing horse motions, such as ``a horse is running/walking/jumping/eating''\footnote{The full list of prompts are included in the supplementary material.}.
We generate $20$ 4D instances from 4D-fy, and $20$ from our method (without text condition).
Note that it takes $12$ hours to generate one 4D-fy instance on one GPU, whereas our model generates 4D animations within a few seconds in a single forward pass.
We first compute the Motion Strength to assess the motion magnitude of the generated videos. We use Flowformer~\cite{huang2022flowformer} to estimate optical flow strengths between consecutive frames of a generated video, and then compute the average of the largest $5$\% optical flows as the Motion Strength.
We present them in random pairs side by side to $33$ participants, and ask them to select one that shows ``a more plausible 3D horse motion sequence''.
As reported in \Cref{tab:cmp_4dfy}, users preferred the 4D instances generated by our method over 4D-fy $83.0$\% of the time.
We show a visual comparison in \Cref{fig:compare_4dfy}.
Notably, 4D-fy produces nearly static animals without perceptible motions despite heavy prompt engineering, whereas our method generates much more plausible motion sequences.

\vspace{-1.3em}
\subsubsection{Quantitative Evaluation.}
Further assessing the quality of the generated 3D motions quantitatively is difficult due to the lack of (1) ground-truth measurements of 3D animal motions, and (2) robust evaluation metrics for generative models.
To evaluate and compare different variants of our model, we design a new metric, bi-directional Motion Chamfer Distance (\textbf{MCD}), computed between a set of generated motion sequences projected to 2D image space and a set of 2D keypoint sequences annotated from videos in APT-36K~\cite{yang2022apt}.
Since the skeleton automatically discovered by our model is different from the 17 keypoints annotated in APT-36K, we first perform 3D reconstruction on all the images in APT-36K, and optimize a linear transformation that maps the 2D projections of the predicted 3D joints to the annotated 2D keypoints following~\cite{kanazawa18cmr}.
To compute MCD, we generate $1,400$ random motion sequences by sampling from the learned motion VAE, each consisting of $10$ frames of 3D articulated poses.
We then project these generated 3D poses to 2D using the viewpoints estimated from APT-36K, and apply the previously optimized transformation to align with the annotated keypoints.
For each annotated keypoint \emph{sequence} in the test set, we find the closest generated motion \emph{sequence} measured by keypoint MSE averaged across all frames, and vice versa for each generated sequence.
We then compute MCD based on the MSE between the closest sequence pairs.
In essence, MCD measures the fidelity of generated motions by comparing the sampled distribution to that of the real motion sequences annotated from videos.
\Cref{tab:motion_gen} compares the results of our final model with two ablated variants.

\begin{table}[t!]
\setlength{\tabcolsep}{0.2cm}
\centering
\footnotesize
\caption{
Motion Chamfer Distance (MCD) on APT-36K~\cite{yang2022apt} for Motion Generation Evaluation. MP: Magicpony, AM: AnimalMotion dataset, TS: temporal smoothness.}
\begin{tabular}{lcccc}
\toprule
Experiment & MCD $\downarrow$ \\
\midrule 
MP + VAE           & 38.77 \\
MP + VAE + AM      & 38.12 \\
MP + VAE + AM + TS (final) & \textbf{38.03} \\

\bottomrule
\end{tabular}
\vspace{-1em}
\label{tab:motion_gen}
\end{table}

\vspace{-0.7em}
\subsection{Single-Image 3D Reconstruction}
\vspace{-0.2em}
We also quantitatively evaluate the monocular 3D reconstruction results of our model and compare with existing methods~\cite{kulkarni19canonical, Li2020umr, kulkarni2020articulation, wu2023magicpony}.
For this purpose, we use PASCAL~\cite{everingham2015pascal}, a widely used benchmarking dataset for 3D reconstruction, as well as the aforementioned APT-36K~\cite{yang2022apt} dataset, both of which come with 2D keypoint annotations.
We compute the commonly used keypoint transfer metric measured by Percentage of Correct Keypoints (\textbf{PCK})~\cite{kanazawa18cmr, Li2020umr, wu2023magicpony}.
Specifically, given a set of annotated visible 2D keypoints on a source image, we identify the closest vertices on the reconstructed 3D mesh, and then project those 3D vertices onto the target 2D image.
We calculate the percentage of the re-projected keypoints that land within a small distance from the annotated keypoints in the target image.
This margin is set to be $0.1$ of the image size following prior work~\cite{kanazawa18cmr, Li2020umr, wu2023magicpony}.
Another commonly used metric is Mask Intersection over Union (\textbf{MIoU}) between the rendered and ground-truth masks, which measures the reconstruction quality in terms of projected 2D silhouettes.
In addition, since APT-36K~\cite{yang2022apt} provides keypoint annotations on video sequences, we also measure the temporal consistency across the reconstructions along the video sequences using a \textbf{Velocity Error},
computed as 
$
\frac{1}{T} \sum_{t=1}^T \| \hat{\delta}_t - \delta_t \| / \delta_t
$,
where $\hat{\delta}_t$ and $\delta_t$ are the keypoint displacements between consecutive frame for predicted and GT pose sequences respectively.
As the predicted poses are different from the GT keypoints, we use the same procedure described in \Cref{sec:motion_eval} to optimize a linear mapping from the predicted poses to the GT keypoints for each method.

\begin{table}[t!]
\setlength{\tabcolsep}{0.2cm}
\small
\centering
\caption{
Comparison of Monocular 3D Reconstruction Results with Different Methods on PASCAL~\cite{everingham2015pascal} and APT-36K~\cite{yang2022apt}.
Our method achieves superior reconstruction accuracy compared to the existing methods, including the recent MagicPony baseline~\cite{wu2023magicpony}.
}
\vspace{-0.5em}
\begin{tabular}{lccccc}
\toprule
      &     \multicolumn{2}{c}{PASCAL~\cite{everingham2015pascal}} & \multicolumn{2}{c}{APT-36K~\cite{yang2022apt}} \\
      \cmidrule(lr){2-3} \cmidrule(lr){4-5}
Method          & PCK $\uparrow$    & Mask IoU $\uparrow$ & PCK $\uparrow$    & Vel. Err. $\downarrow$ \\
\midrule
CSM~\cite{kulkarni19canonical} & 31.2\% & - & - & - \\
UMR~\cite{Li2020umr}                  & 24.4\%	    & - & - & - \\
A-CSM~\cite{kulkarni2020articulation}                 &32.9\%	    & - & - & - \\
MagicPony~\cite{wu2023magicpony}    &42.8\%	    & 64.1\%       & 53.9\%       & 57.3\% \\
Ponymation (ours) &\textbf{48.0\%} & \textbf{71.8\%}       & \textbf{59.9\%}  & \textbf{49.1\%}\\
\bottomrule
\end{tabular}
\label{tab:recon}
\vspace{-1em}
\end{table}

The results are summarized in \Cref{tab:recon}.
The results of MagicPony~\cite{wu2023magicpony} are computed using the publicly released code and models, and the results of other baselines are taken from A-CSM~\cite{kulkarni2020articulation}.
Our model outperforms all previous methods.
In particular, compared to the MagicPony baseline, our model achieves considerable improvement by learning from videos instead of individual images.

Additional ablation studies on the architecture design, discussions on limitations and more visualizations are included the supplementary material.

\section{Conclusions}

We have presented a new method for learning generative models of articulated 3D animal motions from raw Internet videos, without relying on any pose annotations or shape templates. 
To this end, we have proposed a video photo-geometric auto-encoding framework that automatically learns to decompose RGB videos into the underlying 3D shape, articulated motion, and object appearance, simply with the objective of re-rendering the videos.
At the core of this pipeline is a transformer-based architecture that effectively extracts the temporal and spatial structure of the video clip into a latent motion VAE, which enables sampling at inference time to generate new 3D motion sequences.
Experimental results show that the proposed method learns a reasonable distribution of 3D animal motions for several animal categories.
This allows us to instantly turn a single 2D image into 4D animations in a fully automatic fashion, enabling promising downstream applications in game design and movie production.

\paragraph{\bf Acknowledgments.} We thank Zizhang Li, Feng Qiu, and Ruining Li for insightful discussions. The work is in part supported by the Stanford Institute for Human-Centered AI (HAI) and Samsung.

\bibliographystyle{splncs04}
\bibliography{ref}

\begin{thebibliography}{10}
\providecommand{\url}[1]{\texttt{#1}}
\providecommand{\urlprefix}{URL }
\providecommand{\doi}[1]{https://doi.org/#1}

\bibitem{deAguiar08}
de~Aguiar, E., Stoll, C., Theobalt, C., Ahmed, N., Seidel, H.P., Thrun, S.: Performance capture from sparse multi-view video. ACM TOG  (2008)

\bibitem{Ahn2018}
Ahn, H., Ha, T., Choi, Y., Yoo, H., Oh, S.: {Text2Action}: Generative adversarial synthesis from language to action. In: ICRA (2018)

\bibitem{Akhter08nrsfm}
Akhter, I., Sheikh, Y., Khan, S., Kanade, T.: Nonrigid structure from motion in trajectory space. In: NeurIPS (2008)

\bibitem{amir2021deep}
Amir, S., Gandelsman, Y., Bagon, S., Dekel, T.: Deep vit features as dense visual descriptors. In: ECCV Workshop on What is Motion For? (2022)

\bibitem{Andriluka2014}
Andriluka, M., Pishchulin, L., Gehler, P., Schiele, B.: 2d human pose estimation: New benchmark and state of the art analysis. In: CVPR (2014)

\bibitem{Badler1975}
Badler, N.: Temporal Scene Analysis: Conceptual Descriptions of Object Movements. Ph.D. thesis, Queensland University of Technology (1975)

\bibitem{Badler1993}
Badler, N.I., Phillips, C.B., Webber, B.L.: {Simulating Humans: Computer Graphics, Animation, and Control}. Oxford University Press (09 1993)

\bibitem{bahmani20234dfy}
Bahmani, S., Skorokhodov, I., Rong, V., Wetzstein, G., Guibas, L., Wonka, P., Tulyakov, S., Park, J.J., Tagliasacchi, A., Lindell, D.B.: {4D-fy}: Text-to-4d generation using hybrid score distillation sampling. In: CVPR (2024)

\bibitem{Bogo16}
Bogo, F., Kanazawa, A., Lassner, C., Gehler, P., Romero, J., Black, M.J.: Keep it {SMPL}: Automatic estimation of {3D} human pose and shape from a single image. In: ECCV (2016)

\bibitem{Bregler00nrsfm}
Bregler, C., Hertzmann, A., Biermann, H.: Recovering non-rigid 3d shape from image streams. In: CVPR (2000)

\bibitem{videoworldsimulators2024}
Brooks, T., Peebles, B., Holmes, C., DePue, W., Guo, Y., Jing, L., Schnurr, D., Taylor, J., Luhman, T., Luhman, E., Ng, C., Wang, R., Ramesh, A.: Video generation models as world simulators (2024), \url{https://openai.com/research/video-generation-models-as-world-simulators}

\bibitem{caron2021dino}
Caron, M., Touvron, H., Misra, I., J\'egou, H., Mairal, J., Bojanowski, P., Joulin, A.: Emerging properties in self-supervised vision transformers. In: ICCV (2021)

\bibitem{Cashman12dolphin}
Cashman, T.J., Fitzgibbon, A.W.: What shape are dolphins? building 3d morphable models from 2d images. IEEE TPAMI  (2012)

\bibitem{chadwick1989layered}
Chadwick, J.E., Haumann, D.R., Parent, R.E.: Layered construction for deformable animated characters. ACM SIGGRAPH Computer Graphics  (1989)

\bibitem{chan2021piGAN}
Chan, E., Monteiro, M., Kellnhofer, P., Wu, J., Wetzstein, G.: {pi-GAN}: Periodic implicit generative adversarial networks for 3d-aware image synthesis. In: CVPR (2021)

\bibitem{Chan2022eg3d}
Chan, E.R., Lin, C.Z., Chan, M.A., Nagano, K., Pan, B., {De Mello}, S., Gallo, O., Guibas, L., Tremblay, J., Khamis, S., Karras, T., Wetzstein, G.: Efficient geometry-aware {3D} generative adversarial networks. In: CVPR (2022)

\bibitem{Dai12}
Dai, Y., Li, H., He, M.: A simple prior-free method for non-rigid structure-from-motion factorization. In: CVPR (2012)

\bibitem{Debevec12}
Debevec, P.: The light stages and their applications to photoreal digital actors. In: SIGGRAPH Asia (2012)

\bibitem{duggal2022tars3D}
Duggal, S., Pathak, D.: Topologically-aware deformation fields for single-view 3d reconstruction. CVPR  (2022)

\bibitem{everingham2015pascal}
Everingham, M., Eslami, S.A., Van~Gool, L., Williams, C.K., Winn, J., Zisserman, A.: The pascal visual object classes challenge: A retrospective. IJCV  (2015)

\bibitem{gao2022mps}
Gao, X., Yang, J., Kim, J., Peng, S., Liu, Z., Tong, X.: Mps-nerf: Generalizable 3d human rendering from multiview images. IEEE TPAMI  (2022)

\bibitem{Goel20ucmr}
Goel, S., Kanazawa, A., Malik, J.: Shape and viewpoints without keypoints. In: ECCV (2020)

\bibitem{guo2020action2motion}
Guo, C., Zuo, X., Wang, S., Zou, S., Sun, Q., Deng, A., Gong, M., Cheng, L.: Action2motion: Conditioned generation of 3d human motions. In: ACM MM (2020)

\bibitem{habibie2017recurrent}
Habibie, I., Holden, D., Schwarz, J., Yearsley, J., Komura, T.: A recurrent variational autoencoder for human motion synthesis. In: BMVC (2017)

\bibitem{hartley04multiple}
Hartley, R., Zisserman, A.: Multiple View Geometry in Computer Vision. Cambridge University Press, ISBN: 0521540518, second edn. (2004)

\bibitem{he2021challencap}
He, Y., Pang, A., Chen, X., Liang, H., Wu, M., Ma, Y., Xu, L.: {ChallenCap}: Monocular 3d capture of challenging human performances using multi-modal references. In: CVPR (2021)

\bibitem{henter2020moglow}
Henter, G.E., Alexanderson, S., Beskow, J.: {MoGlow}: Probabilistic and controllable motion synthesis using normalising flows. ACM TOG  (2020)

\bibitem{huang2021hierarchical}
Huang, K., Han, Y., Chen, K., Pan, H., Zhao, G., Yi, W., Li, X., Liu, S., Wei, P., Wang, L.: A hierarchical 3d-motion learning framework for animal spontaneous behavior mapping. Nature Communications  (2021)

\bibitem{huang2022flowformer}
Huang, Z., Shi, X., Zhang, C., Wang, Q., Cheung, K.C., Qin, H., Dai, J., Li, H.: {FlowFormer}: A transformer architecture for optical flow. {ECCV}  (2022)

\bibitem{Ionescu2014}
Ionescu, C., Papava, D., Olaru, V., Sminchisescu, C.: {Human3.6M}: Large scale datasets and predictive methods for 3d human sensing in natural environments. IEEE TPAMI  (2014)

\bibitem{jakab24farm3d}
Jakab, T., Li, R., Wu, S., Rupprecht, C., Vedaldi, A.: {Farm3D}: Learning articulated {3D} animals by distilling {2D} diffusion. In: 3DV (2024)

\bibitem{jiang2024motiongpt}
Jiang, B., Chen, X., Liu, W., Yu, J., Yu, G., Chen, T.: {MotionGPT}: Human motion as a foreign language. In: NeurIPS (2024)

\bibitem{Kanazawa_2018_CVPR}
Kanazawa, A., Black, M.J., Jacobs, D.W., Malik, J.: End-to-end recovery of human shape and pose. In: CVPR (2018)

\bibitem{kanazawa18cmr}
Kanazawa, A., Tulsiani, S., Efros, A.A., Malik, J.: Learning category-specific mesh reconstruction from image collections. In: ECCV (2018)

\bibitem{Kanazawa_2019_CVPR}
Kanazawa, A., Zhang, J.Y., Felsen, P., Malik, J.: Learning 3d human dynamics from video. In: CVPR (2019)

\bibitem{kapon2023mas}
Kapon, R., Tevet, G., Cohen-Or, D., Bermano, A.H.: Mas: Multi-view ancestral sampling for 3d motion generation using 2d diffusion. In: CVPR (2024)

\bibitem{kingma2013auto}
Kingma, D.P., Welling, M.: Auto-encoding variational bayes. In: ICLR (2014)

\bibitem{kirillov2019pointrend}
Kirillov, A., Wu, Y., He, K., Girshick, R.: {PointRend}: Image segmentation as rendering. In: CVPR (2020)

\bibitem{kokkinos2021point}
Kokkinos, F., Kokkinos, I.: To the point: Correspondence-driven monocular 3d category reconstruction. In: NeurIPS (2021)

\bibitem{kulkarni2020articulation}
Kulkarni, N., Gupta, A., Fouhey, D.F., Tulsiani, S.: Articulation-aware canonical surface mapping. In: CVPR (2020)

\bibitem{kulkarni19canonical}
Kulkarni, N., Gupta, A., Tulsiani, S.: Canonical surface mapping via geometric cycle consistency. In: ICCV (2019)

\bibitem{Laine2020diffrast}
Laine, S., Hellsten, J., Karras, T., Seol, Y., Lehtinen, J., Aila, T.: Modular primitives for high-performance differentiable rendering. ACM TOG  (2020)

\bibitem{li20online}
Li, X., Liu, S., De~Mello, S., Kim, K., Wang, X., Yang, M., Kautz, J.: Online adaptation for consistent mesh reconstruction in the wild. In: NeurIPS (2020)

\bibitem{Li2020umr}
Li, X., Liu, S., Kim, K., De~Mello, S., Jampani, V., Yang, M.H., Kautz, J.: Self-supervised single-view 3d reconstruction via semantic consistency. In: ECCV (2020)

\bibitem{li2019learning}
Li, Z., Dekel, T., Cole, F., Tucker, R., Snavely, N., Liu, C., Freeman, W.T.: Learning the depths of moving people by watching frozen people. In: CVPR (2019)

\bibitem{Lin2018HumanMM}
Lin, X., Amer, M.R.: Human motion modeling using dvgans. arXiv preprint arXiv:1804.10652  (2018)

\bibitem{ling2023ayg}
Ling, H., Kim, S.W., Torralba, A., Fidler, S., Kreis, K.: Align your gaussians: Text-to-4d with dynamic 3d gaussians and composed diffusion models. In: CVPR (2024)

\bibitem{liu2024lepard}
Liu, D., Stathopoulos, A., Zhangli, Q., Gao, Y., Metaxas, D.: {LEPARD}: Learning explicit part discovery for 3d articulated shape reconstruction. In: NeurIPS (2024)

\bibitem{loper2015smpl}
Loper, M., Mahmood, N., Romero, J., Pons-Moll, G., Black, M.J.: {SMPL}: A skinned multi-person linear model. ACM TOG  (2015)

\bibitem{magnenat1988abstract}
Magnenat-Thalmann, N., Primeau, E., Thalmann, D.: Abstract muscle action procedures for human face animation. The Visual Computer  (1988)

\bibitem{minderer2019unsupervised}
Minderer, M., Sun, C., Villegas, R., Cole, F., Murphy, K.P., Lee, H.: Unsupervised learning of object structure and dynamics from videos. In: NeurIPS (2019)

\bibitem{Muybridge1887horse}
Muybridge, E.: The horse in motion (1887)

\bibitem{newcombe15dynamicfusion}
Newcombe, R.A., Fox, D., Seitz, S.M.: {DynamicFusion}: Reconstruction and tracking of non-rigid scenes in real-time. In: CVPR (2015)

\bibitem{nguyen2019hologan}
Nguyen-Phuoc, T., Li, C., Theis, L., Richardt, C., Yang, Y.L.: {HoloGAN}: Unsupervised learning of 3d representations from natural images. In: ICCV (2019)

\bibitem{Niemeyer2020GIRAFFE}
Niemeyer, M., Geiger, A.: {GIRAFFE}: Representing scenes as compositional generative neural feature fields. In: CVPR (2021)

\bibitem{Niemeyer20DVR}
Niemeyer, M., Mescheder, L., Oechsle, M., Geiger, A.: Differentiable volumetric rendering: Learning implicit 3d representations without 3d supervision. In: CVPR (2020)

\bibitem{chatgpt}
OpenAI: {ChatGPT} (2023), \url{https://chat.openai.com/}

\bibitem{Ormoneit2005}
Ormoneit, D., Black, M., Hastie, T., Kjellström, H.: Representing cyclic human motion using functional analysis. Image and Vision Computing  (2005)

\bibitem{petrovich21actor}
Petrovich, M., Black, M.J., Varol, G.: Action-conditioned 3{D} human motion synthesis with transformer {VAE}. In: ICCV (2021)

\bibitem{petrovich2022temos}
Petrovich, M., Black, M.J., Varol, G.: Temos: Generating diverse human motions from textual descriptions. In: ECCV (2022)

\bibitem{piao2021inverting}
Piao, J., Sun, K., Wang, Q., Lin, K.Y., Li, H.: Inverting generative adversarial renderer for face reconstruction. In: CVPR (2021)

\bibitem{ren2023dreamgaussian4d}
Ren, J., Pan, L., Tang, J., Zhang, C., Cao, A., Zeng, G., Liu, Z.: {DreamGaussian4D}: Generative 4d gaussian splatting. arXiv preprint arXiv:2312.17142  (2023)

\bibitem{pifuSHNMKL19}
Saito, S., Huang, Z., Natsume, R., Morishima, S., Kanazawa, A., Li, H.: {PIFu}: Pixel-aligned implicit function for high-resolution clothed human digitization. In: ICCV (2019)

\bibitem{saito2020pifuhd}
Saito, S., Simon, T., Saragih, J., Joo, H.: {PIFuHD}: Multi-level pixel-aligned implicit function for high-resolution 3d human digitization. In: CVPR (2020)

\bibitem{Schwarz2020graf}
Schwarz, K., Liao, Y., Niemeyer, M., Geiger, A.: {GRAF}: Generative radiance fields for 3d-aware image synthesis. In: NeurIPS (2020)

\bibitem{shen2021dmtet}
Shen, T., Gao, J., Yin, K., Liu, M.Y., Fidler, S.: Deep marching tetrahedra: a hybrid representation for high-resolution 3d shape synthesis. In: NeurIPS (2021)

\bibitem{singer2023makeavideo}
Singer, U., Polyak, A., Hayes, T., Yin, X., An, J., Zhang, S., Hu, Q., Yang, H., Ashual, O., Gafni, O., Parikh, D., Gupta, S., Taigman, Y.: Make-a-video: Text-to-video generation without text-video data. In: ICLR (2023)

\bibitem{sitzmann2019srns}
Sitzmann, V., Zollh{\"o}fer, M., Wetzstein, G.: Scene representation networks: Continuous 3d-structure-aware neural scene representations. In: NeurIPS (2019)

\bibitem{starke2022deepphase}
Starke, S., Mason, I., Komura, T.: {DeepPhase}: Periodic autoencoders for learning motion phase manifolds. ACM TOG  (2022)

\bibitem{stathopoulos2023learning}
Stathopoulos, A., Pavlakos, G., Han, L., Metaxas, D.N.: Learning articulated shape with keypoint pseudo-labels from web images. In: CVPR (2023)

\bibitem{sun2022bkind}
Sun, J.J., Karashchuk, P., Dravid, A., Ryou, S., Fereidooni, S., Tuthill, J., Katsaggelos, A., Brunton, B.W., Gkioxari, G., Kennedy, A., et~al.: {BKinD-3D}: Self-supervised 3d keypoint discovery from multi-view videos. In: CVPR (2023)

\bibitem{bkind2021}
Sun, J.J., Ryou, S., Goldshmid, R., Weissbourd, B., Dabiri, J., Anderson, D.J., Kennedy, A., Yue, Y., Perona, P.: Self-supervised keypoint discovery in behavioral videos. In: CVPR (2022)

\bibitem{sun2022cgof}
Sun, K., Wu, S., Huang, Z., Zhang, N., Wang, Q., Li, H.: Controllable 3d face synthesis with conditional generative occupancy fields. In: NeurIPS (2022)

\bibitem{sun2022cgof++}
Sun, K., Wu, S., Zhang, N., Huang, Z., Wang, Q., Li, H.: Cgof++: Controllable 3d face synthesis with conditional generative occupancy fields. IEEE TPAMI  (2023)

\bibitem{teed2020raft}
Teed, Z., Deng, J.: Raft: Recurrent all-pairs field transforms for optical flow. In: ECCV (2020)

\bibitem{Urtasun2007}
Urtasun, R., Fleet, D.J., Lawrence, N.D.: Modeling human locomotion with topologically constrained latent variable models. In: Elgammal, A., Rosenhahn, B., Klette, R. (eds.) Human Motion -- Understanding, Modeling, Capture and Animation. pp. 104--118. Springer Berlin Heidelberg, Berlin, Heidelberg (2007)

\bibitem{transformer}
Vaswani, A., Shazeer, N., Parmar, N., Uszkoreit, J., Jones, L., Gomez, A.N., Kaiser, {\L}., Polosukhin, I.: Attention is all you need. In: NeurIPS (2017)

\bibitem{wang2017predrnn}
Wang, Y., Long, M., Wang, J., Gao, Z., Yu, P.S.: Predrnn: Recurrent neural networks for predictive learning using spatiotemporal lstms. In: NeurIPS (2017)

\bibitem{wu2021dove}
Wu, S., Jakab, T., Rupprecht, C., Vedaldi, A.: {DOVE}: Learning deformable 3d objects by watching videos. IJCV  (2023)

\bibitem{wu2023magicpony}
Wu, S., Li, R., Jakab, T., Rupprecht, C., Vedaldi, A.: {MagicPony}: Learning articulated 3d animals in the wild. In: CVPR (2023)

\bibitem{wu2021derender}
Wu, S., Makadia, A., Wu, J., Snavely, N., Tucker, R., Kanazawa, A.: De-rendering the world's revolutionary artefacts. In: CVPR (2021)

\bibitem{wu20unsupervised}
Wu, S., Rupprecht, C., Vedaldi, A.: Unsupervised learning of probably symmetric deformable {3D} objects from images in the wild. In: CVPR (2020)

\bibitem{wu2022casa}
Wu, Y., Chen, Z., Liu, S., Ren, Z., Wang, S.: {CASA}: Category-agnostic skeletal animal reconstruction. In: NeurIPS (2022)

\bibitem{Xiao04nrsfm}
Xiao, J., xiang Chai, J., Kanade, T.: A closed-form solution to non-rigid shape and motion recovery. In: ECCV (2004)

\bibitem{xie2023omnicontrol}
Xie, Y., Jampani, V., Zhong, L., Sun, D., Jiang, H.: {OmniControl}: Control any joint at any time for human motion generation. In: ICLR (2024)

\bibitem{yang21lasr}
Yang, G., Sun, D., Jampani, V., Vlasic, D., Cole, F., Chang, H., Ramanan, D., Freeman, W.T., Liu, C.: {LASR}: Learning articulated shape reconstruction from a monocular video. In: CVPR (2021)

\bibitem{yang2021viser}
Yang, G., Sun, D., Jampani, V., Vlasic, D., Cole, F., Liu, C., Ramanan, D.: {ViSER}: Video-specific surface embeddings for articulated 3d shape reconstruction. In: NeurIPS (2021)

\bibitem{yang2022banmo}
Yang, G., Vo, M., Natalia, N., Ramanan, D., Andrea, V., Hanbyul, J.: {BANMo}: Building animatable 3d neural models from many casual videos. In: CVPR (2022)

\bibitem{yang2023rac}
Yang, G., Wang, C., Reddy, N.D., Ramanan, D.: Reconstructing animatable categories from videos. In: CVPR (2023)

\bibitem{yang2022apt}
Yang, Y., Yang, J., Xu, Y., Zhang, J., Lan, L., Tao, D.: {APT-36K}: A large-scale benchmark for animal pose estimation and tracking. In: NeurIPS Dataset and Benchmark Track (2022)

\bibitem{yao2023hi}
Yao, C.H., Hung, W.C., Li, Y., Rubinstein, M., Yang, M.H., Jampani, V.: {Hi-LASSIE}: High-fidelity articulated shape and skeleton discovery from sparse image ensemble. In: CVPR (2023)

\bibitem{yao2022lassie}
Yao, C.H., Hung, W.C., Rubinstein, M., Lee, Y., Jampani, V., Yang, M.H.: {LASSIE}: Learning articulated shape from sparse image ensemble via 3d part discovery. In: NeurIPS (2022)

\bibitem{yao2024artic3d}
Yao, C.H., Raj, A., Hung, W.C., Rubinstein, M., Li, Y., Yang, M.H., Jampani, V.: {ARTIC3D}: Learning robust articulated 3d shapes from noisy web image collections. In: NeurIPS (2024)

\bibitem{zhang2019predicting}
Zhang, J.Y., Felsen, P., Kanazawa, A., Malik, J.: Predicting 3d human dynamics from video. In: ICCV (2019)

\bibitem{zhao2023animate124}
Zhao, Y., Yan, Z., Xie, E., Hong, L., Li, Z., Lee, G.H.: Animate124: Animating one image to 4d dynamic scene. arXiv preprint arXiv:2311.14603  (2023)

\bibitem{zheng2023unified4d}
Zheng, Y., Li, X., Nagano, K., Liu, S., Hilliges, O., De~Mello, S.: A unified approach for text-and image-guided 4d scene generation. In: CVPR (2024)

\bibitem{zhou2023ude}
Zhou, Z., Wang, B.: {UDE}: A unified driving engine for human motion generation. In: CVPR (2023)

\end{thebibliography}

\clearpage

\appendix
{\LARGE\bf Appendices}

\section{Additional Qualitative Results}

\subsection{Additional Motion Generation Results}
\begin{figure*}[t]
\centering
\includegraphics[trim={10pt 0 0 0}, clip, width=0.99\linewidth]{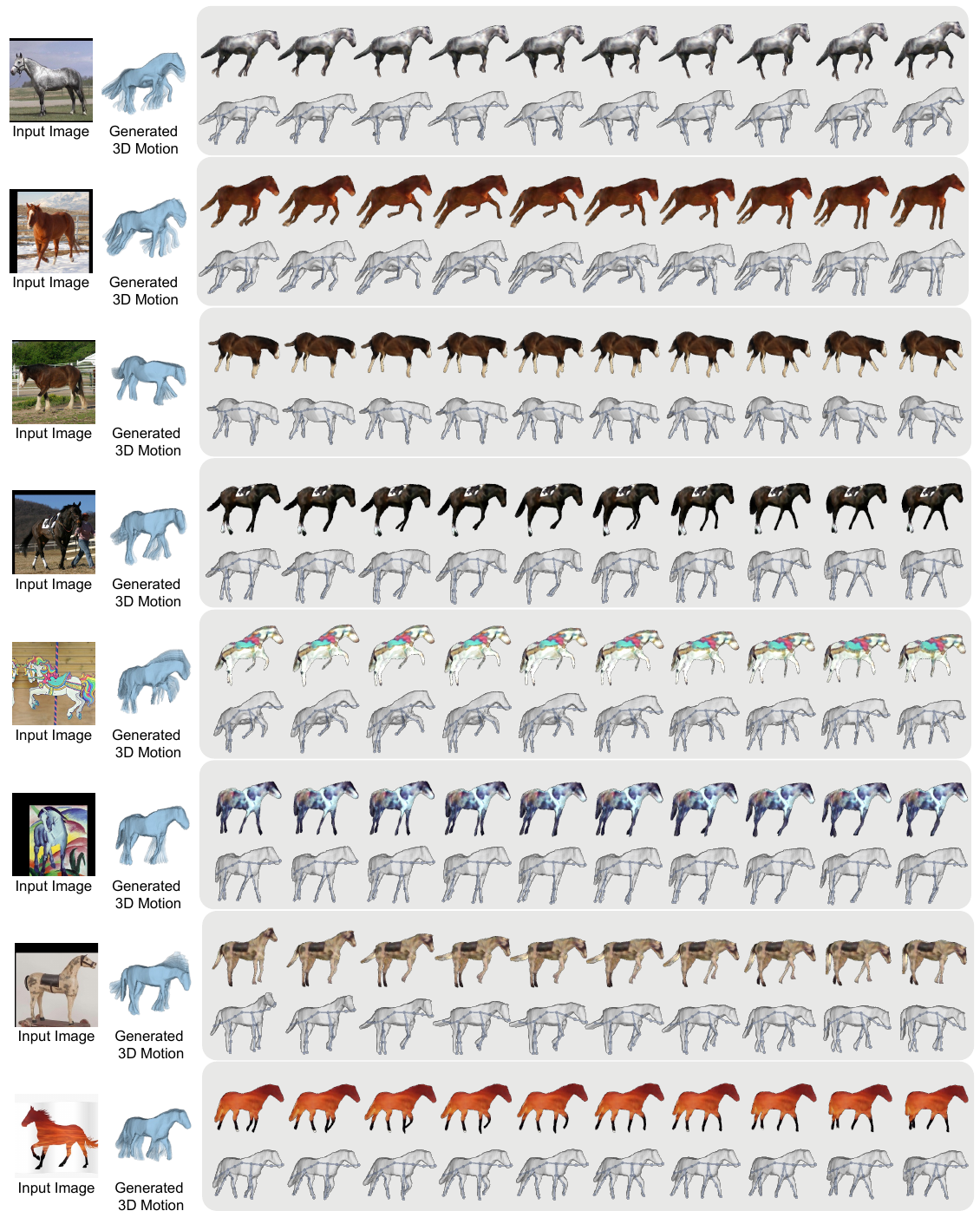}
\caption{\textbf{Additional Motion Generation Results on Horses.}
Conditioned on an input image, which can be either a real photo or a painting of a horse, our model can generate realistic 4D animations of the instance. See the supplementary video for better visualizations. 
}%
\label{fig:supp:horse}
\end{figure*}

\begin{figure*}[t]
\centering
\includegraphics[trim={0 0 0 0}, clip, width=0.99\linewidth]{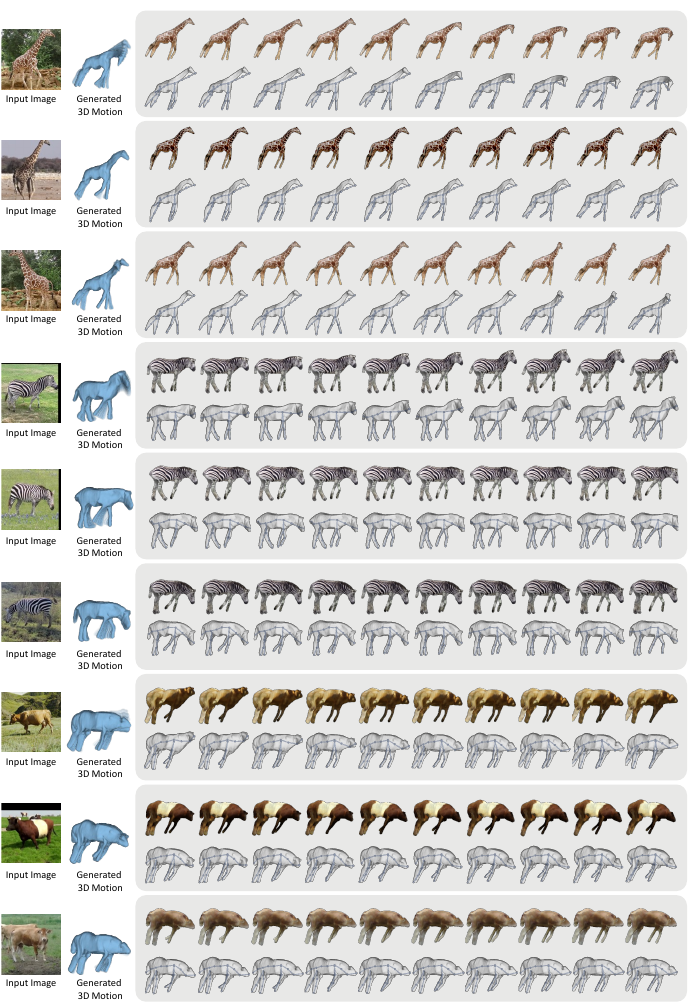}
\caption{\textbf{Additional Motion Generation Results for Other Categories.}
Our model can also be trained on other categories besides horses, and generates realistic motion sequences.
}%
\label{fig:supp:others}
\end{figure*}

Additional generated 3D motion sequences for are shown in \Cref{fig:supp:horse,fig:supp:others}.
Please refer to the video\footnote{\url{https://youtu.be/poc7c-9hCvQ?si=3k874zHackOre94R}} for more 3D animation visualizations.
As shown in the video, by sampling the learned motion latent VAE, we can generate diverse motion patterns, such as \texttt{eating} with the head bending towards the ground, \texttt{walking} with the legs moving alternately, and \texttt{jumping} with the front legs lifted up.

We trained our VAE model with a sequence length of 10 frames.
To produce longer motion sequences as demonstrated in the video, we first sample $2$ latent codes to generate $2$ motion sequences, each comprising 10 frames.
We then optimize $1$ additional transition motion latents by encouraging the poses of the first frame and the last frame to be consistent with the last frame and the first frame of two consecutive sequences previously generated.

\subsection{Qualitative Comparison of Video Reconstruction Results}
\begin{figure}[ht]
\begin{center}
\includegraphics[trim={0 0 0 0}, clip, width=1\linewidth]{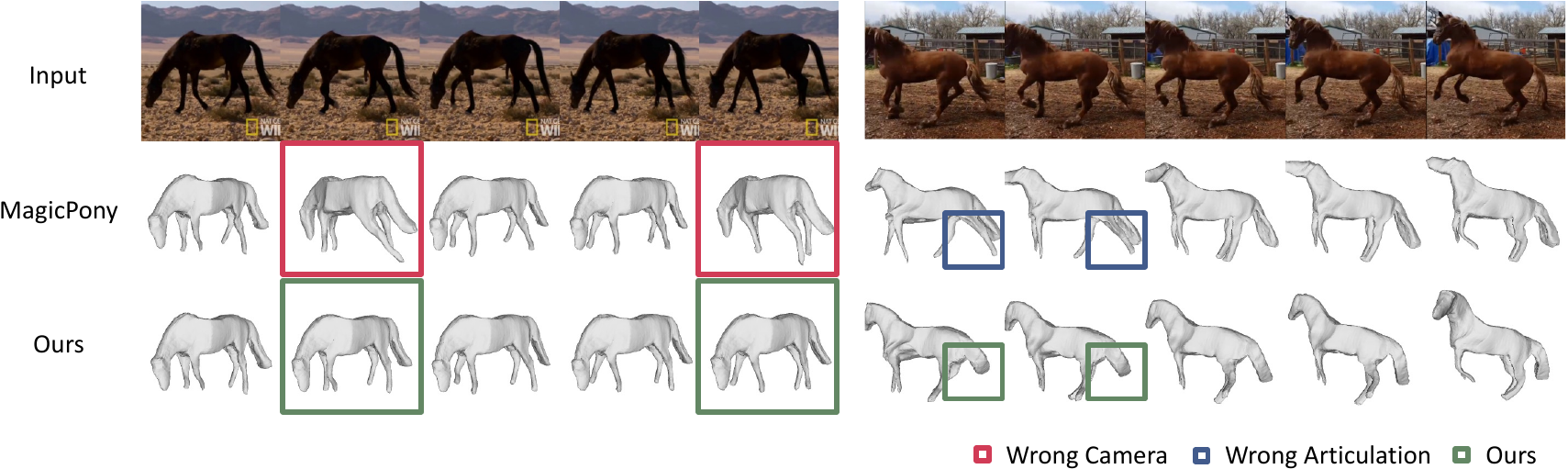}
\end{center}
\vspace{-2em}
\caption{
\textbf{Comparison of 3D Reconstruction Results with MagicPony~\cite{wu2023magicpony}. 
}
With the video training framework, our method produces temporally coherent and more accurate pose predictions. In comparison, the baseline model of MagicPony often predicts incorrect rigid poses $\hat{\xi}_{t,1}$ (red boxes), and incorrect bone articulation $\hat{\xi}_{t,2:B}$ (blue boxes), resulting in inaccurate 3D reconstruction. 
}
\label{fig:smooth}
\vspace{-0.1cm}
\end{figure}

\Cref{fig:smooth} compares the 3D reconstruction results on video sequences obtained from the MagicPony~\cite{wu2023magicpony} model and our proposed method.
Although MagicPony predicts a plausible 3D shape in most cases, it tends to produce temporally inconsistent poses,  including both the rigid pose $\hat{\xi}_{t,1}$ and bone rotations $\hat{\xi}_{t,2:B}$, as highlighted in \Cref{fig:smooth}.
In contrast, our method leverages the temporal signals in training videos, and produces temporally coherent reconstruction results.

\section{Additional Ablation Studies}

\begin{table}[t]
\setlength{\tabcolsep}{0.2cm}
\small
\centering
\caption{Ablation study on the architecture of the motion VAE model.
}
\begin{tabular}{clcc}
\toprule
\multicolumn{1}{c}{} Row & Method   & PCK@0.1  & Mask IoU \\
\midrule
1 & Final (with ST-Transformer) & 37.6\%   & 62.0\% \\
2 & without spatial Transformers $E_\text{s}$, $D_\text{s}$  & 33.4\% & 58.9\% \\
3 & without Teacher Loss $\mathcal{L}_\text{teacher}$ & 32.4\% & 57.9\% \\
4 & without motion VAE      &  44.3\%     & 66.7\% \\

\bottomrule
\end{tabular}

\label{tab:ablation:sttransformer}
\end{table}

\subsection{Spatio-Temporal Transformer Architecture}
We conduct an ablation study to verify the effectiveness of the proposed spatio-temporal transformer architecture.
In particular, we remove each individual component from the final model or replace it with a default option, train the model on the same dataset, and evaluate its performance on 3D reconstruction with the same protocol described in Section 4.3 of the main paper.

First, we remove the spatial transformer encoder and decoder, $E_\text{s}$ and $D_\text{s}$, and report the results in row 2 of \Cref{tab:ablation:sttransformer}.
In this variant, specifically, instead of using the spatial transformer encoder $E_\text{s}$ to fuse bone-specific local image features before passing them to the temporal transformer encoder $E_\text{t}$, we directly feed the global image features $\{\phi_1, \cdots, \phi_T\}$ into the temporal encoder.
Similarly, we also remove the spatial decoder $D_\text{s}$, and directly decode a fixed set of bone rotations from the temporal transformer decoder $D_\text{t}$.

Compared to the final model with spatio-temporal transformer architectures in row 1 of \Cref{tab:ablation:sttransformer}, the variant without spatial transformer results in less accurate reconstructions, and hence lower scores on the metrics.
This confirms the effectiveness of the proposed spatial transformer in extracting motion-specific spatial information from the images.

\subsection{Teacher Loss}
We also demonstrate the effect of the Teacher Loss $\mathcal{L}_\text{teacher}$ introduced in Section 3.3 of the main paper.
We train a variant motion VAE model without this loss, and report its reconstruction performance in Row $3$ of ~\Cref{tab:ablation:sttransformer}.
Without $\mathcal{L}_\text{teacher}$, the model fails to learn accurate poses effectively, leading to degraded reconstruction results.
This is mainly because that training the motion VAE from scratch is computationally inefficient with an expensive rendering step in the loop, and the Teacher Loss can significantly improve training efficiency.

\begin{table}[t!]
\setlength{\tabcolsep}{0.2cm}
\centering
\footnotesize
\caption{
Ablation study with different sequence lengths for motion generation evaluated using Motion Chamfer Distance (MCD) on APT-36K~\cite{yang2022apt}.}
\begin{tabular}{ccccc}
\toprule
Sequence Length & $K=10$ & $K=20$ & $K=50$ \\
\midrule
MCD $\downarrow$ & \textbf{38.03} & 38.25 & 39.25 \\
\bottomrule
\end{tabular}

\label{tab:supp_fid}
\end{table}

\subsection{Sequence Length.}
We conducted experiments to understand the effect of different sequence lengths during training ($K=10, 20, 50$ frames).
For a fair comparison, to evaluate the longer motion sequences generated by these variants ($K=20, 50$), we divide them into consecutive sub-sequences of $10$ frames, and average the MCD metric across the subsequences.
We use the same metric as introduced in Section 4.2 of the main paper, the Motion Chamfer Distance (MCD) calculated between generated sequences and the annotated sequences in the APT-36K dataset~\cite{yang2022apt}.
The results are presented in \Cref{tab:supp_fid}.

Upon analyzing the results, we observed that the generated sequences still look plausible as the sequence length increases from $10$ to $20$.
However, a notable degradation in quality is observed as the sequence length increases to $50$.
This could potentially be attributed to the limited capacity of the motion VAE model as well as the limited size of the training dataset.
For our final model, we set the sequence length to $10$, which tends to yield the most satisfactory results with a reasonable training efficiency.

\subsection{KL Loss Weight.}
To train the motion VAE, in addition to the reconstruction losses, we also use the Kullback–Leibler (KL) divergence loss $\mathcal{L}_\text{KL}$ in Equation (6) in the main paper.
We conducted an ablation study on its weight $\lambda_\text{KL}$ to assess its impact on the overall 3D reconstruction accuracy.
As shown in \Cref{tab:ablation:weights}, $\lambda_\text{KL}=0.001$ achieves the best reconstruction results, and is used in all
experiments in the main paper.

\begin{table}[t!]
\setlength{\tabcolsep}{0.2cm}
\small
\centering
\caption{Ablation study on the weight of the KL divergence loss $\lambda{L}_\text{KL}$.}
\begin{tabular}{lcc}
\toprule
& PCK@0.1 & Mask IoU \\
\midrule
$\lambda_\text{KL} = 0.01$      &  33.58\%     & 59.85\%   \\
$\lambda_\text{KL} = 0.001$     & \textbf{37.63\%}  & \textbf{62.03\%}  \\
$\lambda_\text{KL} = 0.0001$    &  35.75\%     & 61.11\% \\

\bottomrule
\end{tabular}

\label{tab:ablation:weights}
\end{table}

\begin{figure*}[t]
\centering
\includegraphics[trim={0 0 0 0}, clip, width=0.99\linewidth]{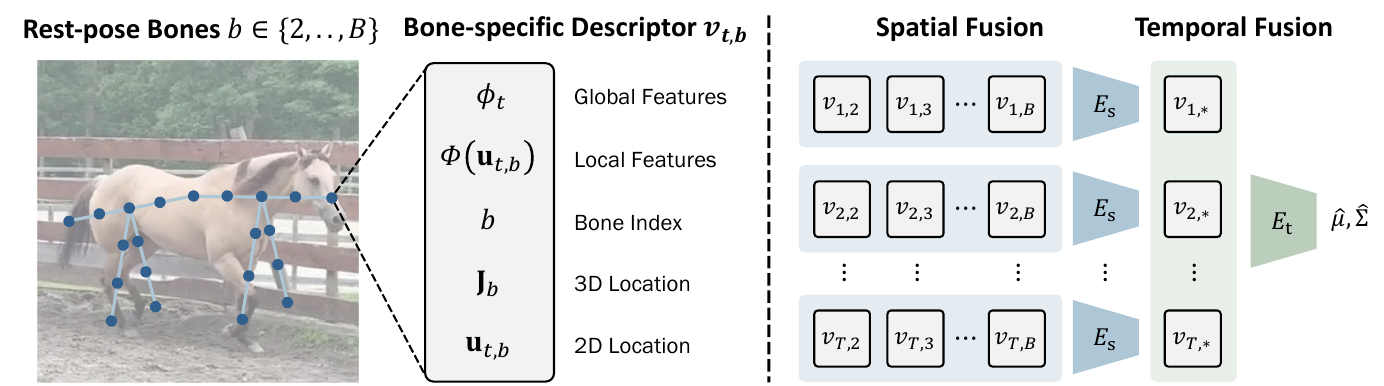}
\caption{\textbf{Illustration of the Spatio-temporal Transformer-based Motion Encoder.} For each frame, the bone-specific features $\{\nu_{t,b}\}_{b=2}^B$ are first extracted from image features and fused by a spatial encoder $E_\text{s}$ to obtain a single feature vector $\nu_{t,*}$. A temporal encoder $E_\text{t}$ then further fuses the feature vectors of all frames $\{\nu_{t,*}\}_{t=1}^T$ and produces the motion VAE distribution parameters $\hat{\mu}$ and $\hat{\Sigma}$. Please refer to the Section~3.2 in the main paper for detail.
}%
\label{fig:supp:arch}
\end{figure*}

\section{Additional Technical Details}

\subsection{Architecture Details}
As explained in the paper, we adopt a spatio-temporal transformer architecture for sequence feature encoding and motion decoding. For better illustrating the architecture, we depict the framework of the spatial and temporal transformer encoders in \Cref{fig:supp:arch}.
Also, as presented in \Cref{tab:arch_st_trans}, we use the $4$-layer transformer to implement the spatial and temporal transformer encoders $E_\text{s}$, $E_\text{t}$ and decoders $D_\text{s}$, $D_\text{t}$.
Given the DINO features of the input image, we first concatenate the bone position as Positional Encoding to obtain the bone-specific feature descriptors $\nu_{t,b}$ with shape (BoneNum, FrameNum, FeatureDim) = ($20 \times 10 \times 640$).
Then we map the feature dimension to $256$ with a simple Linear layer, and concatenate an additional BoneFeatureQuery token.
We use the $4$-layer transformer $E_\text{s}$ to aggregate all the bone-specific feature descriptors into a per-frame pose feature $\nu_{t,*}$, and subsequently $E_\text{t}$ to aggregate all frame-specific features into the VAE distribution parameters, including the mean $\hat{\mu}$ and variance $\hat{\Sigma}$.
Using the reparametrization trick, we then sample a latent code $z$ from the Gaussian distribution $z\sim \mathcal{N}(\hat{\mu}, \hat{\Sigma})$, which is first decoded by the temporal decoder $D_\text{t}$ and the spatial decoder $D_\text{s}$ into a final sequence of bone rotation angles $\hat{\xi}_{*,2:B} \in \mathbb{R}^{20 \times 10 \times 3}$.

\begin{table}[t]
\setlength{\tabcolsep}{0.2cm}
\small
\begin{center}
\caption{Architecture of the proposed spatio-temporal transformer VAE.
}
\begin{tabular}{lc}
\toprule
  Operation & Output Size \\ \midrule
  Positional Encoding & 20 $\times$ 10 $\times$ 640\\
 Linear(640, 256) & 20 $\times$ 10 $\times$ 256\\
 Concat BoneFeatQuery & 21 $\times$ 10 $\times$ 256\\
 TransformerLayer $\times$ 4 & 1 $\times$ 10 $\times$ 256\\
 Reshape & 10 $\times$  1 $\times$ 256\\
 Concat muQuery and sigmaQuery & 12 $\times$ 1 $\times$ 256\\ 
Positional Encoding & 12 $\times$ 1 $\times$ 256\\
 TransformerLayer $\times$ 4 & 2 $\times$ 1 $\times$ 256\\ 
 \midrule
 Reparameterizion & 1 $\times$ 1 $\times$ 256\\
 \midrule
 TransformerLayer $\times$ 4 & 10 $\times$ 1 $\times$ 256\\
 Reshape & 1 $\times$ 10 $\times$ 256\\
 TransformerLayer $\times$ 4 & 20 $\times$ 10 $\times$ 256\\
 Linear(256, 3) & 20 $\times$ 10 $\times$ 3\\
\bottomrule
\end{tabular}

\label{tab:arch_st_trans}
\end{center}
\end{table}

\subsection{Articulation Model Specifications}
The configuration of bone topology and skinning weights was established following Magicpony~\cite{wu2023magicpony}.
Here, we give a brief recap of the model.

\subsubsection{Posed Shape.}

The blend skinning model for posing~\cite{magnenat1988abstract,chadwick1989layered,wu2023magicpony} was utilized to articulate the skeleton into a specific pose. This model is parameterised by $B-1$ bone rotations $\xi_b \in SO(3), b=2,\dots,B$, and the viewpoint $\xi_1 \in SE(3)$. A set of rest-pose joint locations $\mathbf{J}_b$ was initialized on the instance mesh using straightforward heuristics. Each bone $b$, excluding the root, has a single parent $\pi(b)$, thereby forming a tree structure.

Each vertex $V_i$ is linked to the bones via the skinning weights $w_{ib}$, determined based on their relative proximity to each bone. The vertices are then posed using the linear blend \emph{skinning equation}:
\begin{equation}\label{e:skinning}
\begin{aligned}
    V_i(\xi) =\left( \sum_{b=1}^B w_{ib} G_b(\xi) G_b(\xi^*)^{-1} \right) V_{\text{ins},i}, \\
    G_1 = g_1,
    ~~
    G_b = G_{\pi(b)} \circ g_b,
    ~~
    g_b(\xi) = \begin{bmatrix}
        R_{\xi_b} & \mathbf{J}_b \\ 0 & 1 \\
    \end{bmatrix},
\end{aligned}
\end{equation}
where $\xi^*$ denotes the bone rotations at the rest pose.

\subsubsection{Bone Topology} 
For all quadrupedal animals examined in this paper, a chain of $8$ bones of equal lengths was estimated. These bones lie on two line segments that extend from the centre (root) of the rest-pose mesh to the two most extreme vertices along the $z$-axis ($4$ bones on each side), thereby forming a ``spine''. Then the root joint was slightly elevated, and $4$ sets of bones were added to model the legs. The foot joints were first identified as the lowest points of the mesh (in the $y$-axis) in each of the four $xz$-quadrants. Subsequently, $4$ line segments were drawn from the foot joints to their nearest spine joints, and a chain of $3$ bones of equal lengths was defined on each of the segments, representing each leg.

\subsubsection{Skinning Weight} 
The skinning weight $w_{i,b}$, which associates each vertex $V_{\text{ins},i}$ with the bones, was defined as follows:
\begin{equation}\label{e:skinning_w}
\begin{aligned}
  w_{i,b} &= \frac{e^{-d_{i,b}/\tau_\text{s}}}{\sum_{k=1}^{B}{e^{-d_{i,k}/\tau_\text{s}}}},
  \\
  \text{where} \quad d_{i,b} &= \min_{r \in [0,1]} \|V_{\text{ins},i} - r\tilde{\mathbf{J}}_b - (1-r)  \tilde{\mathbf{J}}_{\pi(b)}\|^2_2
\end{aligned}
\end{equation}
In this context, $d_{i,b}$ is the minimal distance from the vertex $V_{\text{ins},i}$ to each bone $b$, defined by the rest-pose joint locations $\tilde{\mathbf{J}}_b$ and $\tilde{\mathbf{J}}_{\pi(b)}$ in world coordinates. $\tilde{\mathbf{J}}_{\pi(b)}$ denotes the parent joint of $\tilde{\mathbf{J}}_{b}$. The temperature parameter $\tau_\text{s}$ is set to $0.5$.

\subsection{Text Prompts for 4D-fy Evaluation}
We provide the 4D-fy~\cite{bahmani20234dfy} model with a list of text prompts, which are enriched by ChatGPT~\cite{chatgpt} from a list of basic prompts describing horse motions.
The complete list is enumerated in the following:

\begin{itemize}
    \item A horse is running.
    \item A horse is running.
    \item A majestic horse galloping swiftly across the verdant meadow.
    \item An energetic steed dashing with unbridled enthusiasm under the azure sky.
    \item A spirited horse racing with the wind, its mane flowing like waves.
    \item A horse is walking.
    \item A horse is walking.
    \item A serene horse ambling gently through a misty forest at dawn.
    \item An elegant steed strolling leisurely along a cobblestone path.
    \item A calm equine sauntering with grace across a blooming meadow.
    \item A horse is eating.
    \item A horse is eating.
    \item A serene horse gently nibbling on the lush green grass of a tranquil meadow.
    \item An elegant equine gracefully bending to graze on the dew-kissed clover.
    \item A peaceful steed leisurely munching on hay in the golden light of dawn.
    \item A horse is jumping.
    \item A horse is jumping.
    \item A majestic horse soaring effortlessly over a rustic wooden fence, its muscles rippling with power.
    \item An agile steed leaping gracefully, silhouetted against the vibrant hues of the setting sun.
    \item A spirited equine vaulting energetically over an obstacle, mane flowing like a river in the wind.
    
\end{itemize}

\vspace{-0.1cm}

\section{Limitations and Future Directions}
While the model demonstrates promising results, there are several areas where further improvements can be made.

A significant limitation is that the articulated motions are learned on top of a fixed bone topology, which is pre-defined using strong heuristics, such as the number of legs. This approach may not effectively generalize across diverse animal species. A potential avenue for future research could involve the joint discovery of the articulation structure in conjunction with video training.

Additionally, the current model does not distinguish between different legs due to the nature of the DINO features. This can result in a ``curious legs'' problem, where the model confuses left and right legs of an animal seen from the side.
This can be observed in the reconstruction results and subsequently in the generated motion sequences, and is also a common issue even with the most powerful video generation models~\cite{videoworldsimulators2024}.
Accurately capturing the leg ordering and precise motion is an intriguing challenge for future research in motion generation.

\section{Societal Impact}

The task of generating 3D motion from unlabeled videos represents a fundamental challenge in the fields of computer vision and computer graphics, in order to extend our current models to the long tail distribution of all kinds of objects in the real world.
As an initial exploration in this area, our aim is to stimulate increasing interest and research in this direction.
The continued advancement in this field holds great potential of significantly improving the diversity and quality of 3D and 4D models of real-world objects, thereby supporting numerous following applications in virtual reality, robotics and scientific discovery.

\end{document}